\documentclass[manuscript,NETN]{stjour}
 \journalname{Network Neuroscience}
 \usepackage{todonotes}
 \usepackage{graphicx}

\graphicspath{{figures/}}

\newcommand{\vc}[1]{{\bf #1}}

\newcommand{\ma}[1]{{\bf #1}}

\articletype{Methods}

\supportinginfo{}

\received{}
\accepted{}
\published{}
\setdoi{} 

\handlingeditor{}

\begin{document}
\title[]{Guided Graph Spectral Embedding}
\subtitle{Application to the {C.~elegans} Connectome}


\author[Author Names]
{Miljan Petrovic\affil{1,2}, Thomas A.W. Bolton\affil{1,2}, Maria Giulia Preti\affil{1,2}, Rapha\"el Li\'egeois\affil{1,2} and Dimitri Van De Ville\affil{1,2}}

\affiliation{1}{Institute of Bioengineering, \'{E}cole Polytechnique F\'ed\'erale de Lausanne, Campus Biotech, Geneva, Switzerland}
\affiliation{2}{Department of Radiology and Medical Informatics, University of Geneva, Geneva, Switzerland}

\correspondingauthor{Dimitri Van De Ville}{dimitri.vandeville@epfl.ch}

\keywords{spectral graph domain, graph embedding, low-dimensional space, focused connectomics}

\begin{abstract}
Graph spectral  analysis can yield meaningful embeddings of graphs by providing insight into distributed features not directly accessible in nodal domain. Recent efforts in graph signal processing have proposed new decompositions---\textit{e.g.}, based on wavelets and Slepians---that can be applied to filter signals defined on the graph. In this work, we take inspiration from these constructions to define a new guided spectral embedding that combines maximizing energy concentration with minimizing modified embedded distance for a given importance weighting of the nodes. We show that these optimization goals are intrinsically opposite, leading to a well-defined and stable spectral decomposition. The importance weighting allows to put the focus on particular nodes and tune the trade-off between global and local effects. Following the derivation of our new optimization criterion, we exemplify the methodology on the \textit{C. elegans} structural connectome. The results of our analyses confirm known observations on the nematode's neural network in terms of functionality and importance of cells. Compared to Laplacian embedding, the guided approach, focused on a certain class of cells (sensory neurons, interneurons or motoneurons), provides more biological insights, such as the distinction between somatic positions of cells, and their involvement in low or high order processing functions.
\end{abstract}


\section{Introduction}
Many aspects of network science relate to graph partitioning---the grouping of nodes in subgraphs---and graph embedding---their representation in a low-dimensional space that accounts for graph topology~\citep{VonLuxburg2007}. Spectral graph theory motivates analytical methods based on the eigenvectors of fundamental graph operators, such as the adjacency and the Laplacian operators~\citep{Chung.1997}. For instance, the well-known graph cut problem can be convexly relaxed and solved by thresholding of the Laplacian eigenvector with the smallest non-zero eigenvalue, known as the Fiedler vector~\citep{Fiedler.1989}. More recently, new approaches in graph signal processing have taken advantage of the Laplacian eigenvectors to define the graph Fourier transform, which can then be used to process (\textit{i.e.}, filter) graph signals in the spectral domain~\citep{Shuman.2013,Ortega2018}; the spectral graph wavelet transform by \cite{Hammond.2011} is one such example. 

The Laplacian eigenvectors also provide a meaningful embedding by mapping nodes onto a line, or higher-dimensional representation, that minimizes distances between connected nodes~\citep{Belkin.2003}. Other well-known embedding techniques use different metrics for distance in order to assess local graph properties, ranging from simple Euclidean distance in locally linear embedding~\citep{Roweis2000}, to shortest path in Isomap~\citep{Tenenbaum2000}, transition probability ~\citep{Shen2008}, or conditional probability of an edge in t-distributed stochastic neighbor embedding~\citep{vanDerMaaten2008}. A time-dependent dynamical similarity measure has also been introduced ~\citep{Schaub2018}. In addition, efforts have been made to employ global properties of the graph, such as in Sammon mapping~\citep{Sammon1969}, where a cost function including all pairwise distances is optimized. In this manner, embedding is performed while taking in consideration both local (neighborhood) and global (distant nodes) properties of the graph. However, these techniques are not aware of the network at the mesoscale: one cannot guide the embedding by giving a certain subgraph more importance while still preserving local features and global topology characteristics.

In essence, the most powerful feature of graph spectral embedding is to effectively summarize local structure across the graph into low-dimensional global patterns. This is achieved, for instance, with the recently introduced concept of graph Slepians; \textit{i.e.}, graph signals that are bandlimited and take into account a subset of selected nodes. Specifically, two types of Slepian designs that respectively optimize for energy concentration and modified embedded distance have been introduced \citep{VDV_Slepians_1,vandeville1701}. 

In this work, we further build on this framework by providing a simple way to guide analyses with additional flexibility. \textit{Guidance} includes the selection of a given subgraph or group of nodes to study, and the ability to specify the intensity of the focus set on these selected nodes. 
With respect to graph Slepians, we hereby provide several extensions. First, we allow the selection process to be weighted, so that the importance of a node can be gradually changed. Second, we propose a new criterion that meaningfully combines the two existing ones; \textit{i.e.}, we want to maximize energy concentration and minimize modified embedded distance at the same time. Third, as we detail below, these two criteria are counteracting, and hence, we obtain stable solutions even at full bandwidth, where the original Slepian designs degenerate numerically. Fourth, we show how this criterion can be rewritten as an eigenvalue problem of an easy modification of the adjacency matrix, which can be interpreted as reweighting paths in the graph, and thus significantly simplifies the whole Slepian concept. The solution of the eigendecomposition then defines the guided spectral domain, spanned by its eigenvectors. We illustrate the proposed approach with a proof-of-concept on the \textit{Caenorhabditis elegans} (\textit{C. elegans}) connectome. Through spectral embedding-based visualization, we observe the effects of focusing on a specific cellular population made of sensory neurons, interneurons or motoneurons, and we reveal trajectories of these neurons as a function of focus strength.

\begin{figure}
\centering
\includegraphics[]{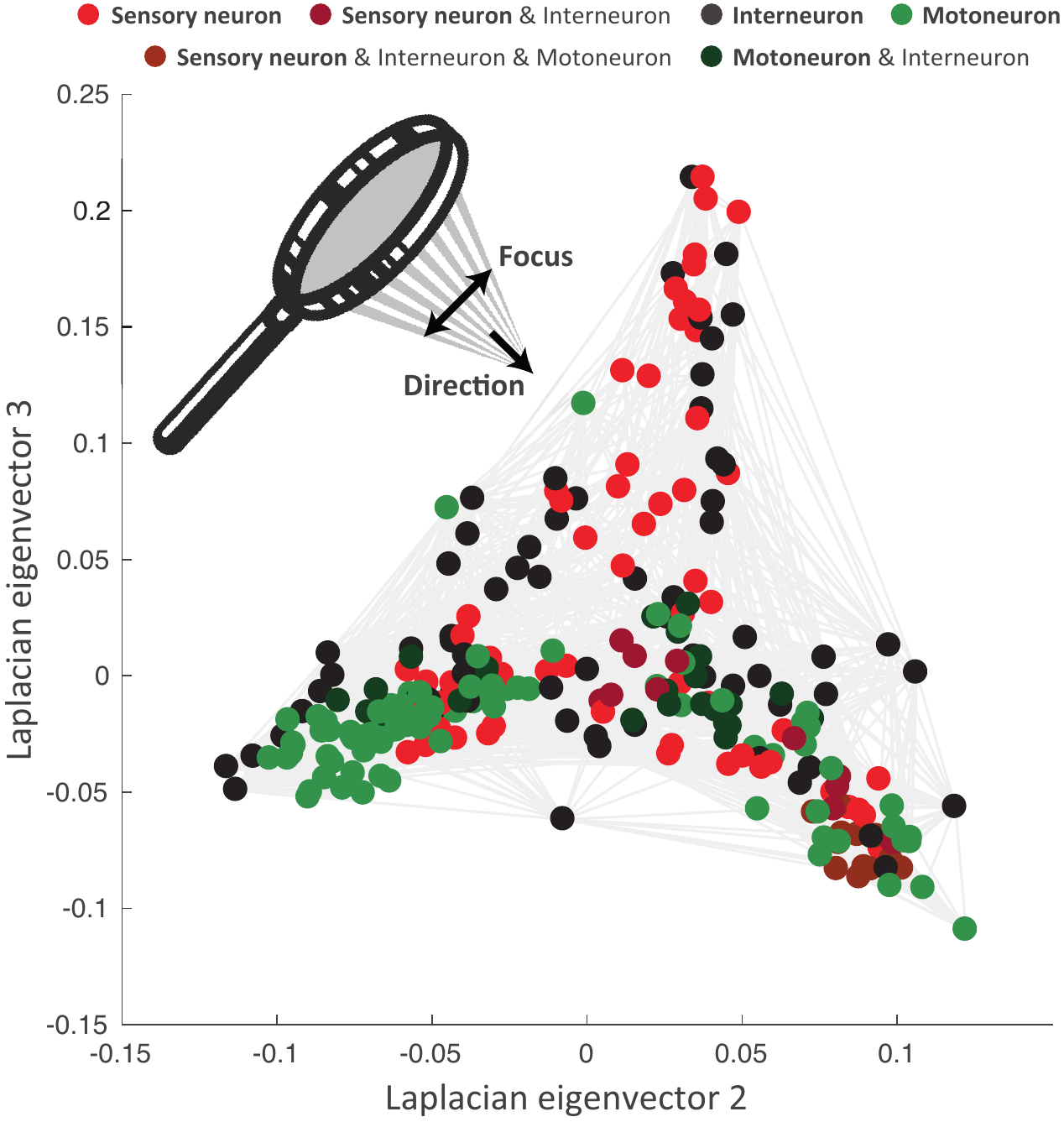}
\caption{\label{fig:principle} Spectral embedding of the \textit{C.~elegans} connectome according to the eigenvectors of the Laplacian matrix with second and third smallest eigenvalues. The purpose of this work is to introduce guided spectral analysis; that is, to indicate direction by selecting a subset of nodes, and to adjust the strength of the focus set on this subset. Each colored circle in the figure depicts one \textit{C. elegans} neuron. Light gray strokes link the cells that are connected by gap junctions or chemical synapses. Labels and connectivity were retrieved from \cite{Varshney2011}.}
\end{figure}

\section{Methods}

\subsection{Essential Graph Concepts}
\label{sec:methods_essential}
We consider an undirected graph with $N$ nodes, labeled $1,2,\ldots,N$. The edge weights are contained in the symmetric weighted adjacency matrix $\tilde{\ma A}$ with non-negative real-valued elements $\tilde{a}_{i,j}$, $i,j=1,\ldots,N$. We also assume that the graph contains no self-loops; \textit{i.e.}, all diagonal elements $\tilde{a}_{i,i}$ are zero.
The degree matrix $\ma D$ is the diagonal matrix with elements $d_{i,i}=\sum_{j=1}^N \tilde{a}_{i,j}$. The graph Laplacian is defined as $\tilde{\ma L}=\ma D-\tilde{\ma A}$ and can be interpreted as a second-order derivative operator on the graph. Here, we use the symmetrically normalized variants of the adjacency $\tilde{\ma A}$ and graph Laplacian $\tilde{\ma L}$ defined as $\ma A=\ma D^{-1/2}\tilde{\ma A}\ma D^{-1/2}$ and $\ma L=\ma I-\ma A$. This normalization is often used in applications to emphasize the changes in topology and not in nodal degree \citep{deLange2014}.

Let us define a graph signal as a vector of length $N$ that associates a value to each node~\citep{Shuman.2013}. One way to recognize the importance of the Laplacian and its eigendecomposition is to consider the smoothness of a graph signal $\vc x$ as

\begin{equation}
  \label{eq:lap_embedding}
  \vc{x}^\top \ma{L} \vc{x} = \sum_{i,j=1}^{N} a_{i,j} (x_i - x_j)^2,
\end{equation}
which sums squared differences between signal values on nodes that are connected, proportionally to their link strength $a_{i,j}$. The eigenvectors of $\ma L$ minimize this distance that is reflected by the eigenvalues, sorted by convention increasingly as ${\lambda_1=0\le \lambda_2 \le \ldots \le \lambda_N}$. Therefore, considering the eigenvectors associated to the smallest non-zero eigenvalues provides the Laplacian embedding of the nodes that minimizes distance in a lower-dimensional space~\citep{Belkin.2003}. The eigenvector with the smallest non-zero eigenvalue is also known as the Fiedler vector~\citep{Fiedler.1989}, which relates to the solution of the convex relaxation of the graph cut problem~\citep{VonLuxburg2007}.

Therefore, the eigendecomposition $\ma L=\ma U\ma\Lambda \ma U^\top$ of the graph Laplacian is the cornerstone of spectral methods for graphs, as the eigenvectors $\{\mathbf{u}_k\}$, $k=1,...,N$ (columns of $\ma U$) play the role of graph Fourier components, and the associated eigenvalues $\{\lambda_k\}$, $k=1,...,N$, of frequencies~\citep{Chung.1997}. The graph Fourier transform (GFT)  then provides the link between a graph signal $\vc x$ and its spectral coefficients given by vector $\hat{\vc x}$:

$$
   \vc x = \ma U \hat{\vc x}, \text{ and } \hat{\vc x} = \ma U^\top \vc x.
$$
%
\subsection{Graph Slepians}

In earlier work, the combination of the concepts of selectivity and bandwidth for graph signals has been used to define ``graph Slepians''~\citep{Tsitsvero.2016,VDV_Slepians_1,vandeville1701}; \textit{i.e.}, bandlimited graph signals with maximal energy concentration in the subset of nodes $\mathcal{S}$---a generalization of prolate spheroidal wave functions that were proposed fifty years ago on regular domains to find a trade-off between temporal and spectral energy concentrations~\citep{Slepian.1961,Slepian.1978}. The presence or absence of a node in $\mathcal{S}$ is encoded by the diagonal elements of the selection matrix $\ma S$; that is, we have $S_{i,i}=\delta_{i\in\mathcal{S}}$, $i=1,\ldots,N$, where $\delta$ is the Kronecker delta.
The  Slepian design then boils down to finding the linear combination of Laplacian eigenvectors, encoded by spectral coefficients $\hat{\vc g}$, within the bandlimit $W$ with maximal energy in $\mathcal{S}$, reverting to the Rayleigh quotient

\begin{equation}
\label{eq:slepian}
  \mu = \frac{\hat{\vc g}^\top \ma{W}^\top \ma{U}^\top \ma{S} \ma{U} \ma W \hat{\vc g}}{\hat{\vc g}^\top \hat{\vc g}},
\end{equation}
where $\ma W$ is a spectral selection matrix that has $W$ ones on its diagonal followed by $N-W$ zeros. This problem can be solved by the eigendecomposition of the concentration matrix $\ma{C}= \ma W^\top \ma{U}^\top \ma{S} \ma{U} \ma W$ as $\ma C\mathbf{\hat{g}}_k=\mu_k \mathbf{\hat{g}}_k$, $k=1,...,W$. The graph Slepians $\mathbf{g}_k=\ma U \mathbf{\hat{g}}_k$, $k=1,...,W$, are orthonormal over the entire graph as well as orthogonal over the subset $\mathcal{S}$; \textit{i.e.}, we have $\mathbf{g}_k^\top \mathbf{g}_l=\delta_{k-l}$ as well as $\mathbf{g}_k^\top \ma{S} \mathbf{g}_l  = \mu_k \delta_{k-l}$.

For the purpose of this work, we introduce the set of bandlimited graph signals

$$
   \mathcal{B}_W = \left\{ \vc x | \hat{\vc x}=\ma W\hat{\vc x} \right\},
$$
such that we can then rewrite the Slepian criterion of Eq.~(\ref{eq:slepian}) directly in the vertex domain as

\begin{equation}
   \mu = \frac{\vc g^\top \ma S \vc g}{\vc g^\top \vc g} \quad\text{s.t. }\vc g \in \mathcal{B}_W.
\end{equation}

An alternative Slepian design was also proposed in \cite{vandeville1701}---see also \cite{Huang_2018}, modifying the Laplacian embedded distance of Eq.~(\ref{eq:lap_embedding}) as follows:

\begin{equation}
\label{XI_EQ}
  \xi = \frac{\vc g^\top \ma L^{1/2} \ma S \ma L^{1/2} \vc g}{\vc g^\top \vc g} \quad\text{s.t. }\vc g \in \mathcal{B}_W.
\end{equation}

The Laplacian embedded distance $\vc{x}^\top \ma{L} \vc{x}$ is a measure of smoothness of the vector $\vc{x}$ over the graph, which is why eigenvectors of $\ma L$ with increasing eigenvalues are ordered according to smoothness. Imposing the modification with the selection matrix $\ma S$ focuses the smoothness on a certain subgraph, notwithstanding how the signal behaves outside it. Eq. (\ref{XI_EQ}) can also be seen as a generalization of Laplacian embedding, since $\ma L^{1/2} \ma S \ma L^{1/2}$ reverts to $\ma L$ for the special case of $\ma S=\ma I$.

It is important to realize that the eigenvalues $\{\mu_k\}$ of the original design reflect the energy concentration in the subset $\mathcal{S}$, while the eigenvalues $\{\xi_k\}$ of the alternative design correspond to a modified embedded distance that can be interpreted as a ``frequency value'' localized in $\mathcal{S}$, in analogy to the global GFT case. Consequently, ``interesting'' eigenvectors correspond to those with high $\mu_k$, concentrated in the subset $\mathcal{S}$, or low $\xi_k$, showing the main localized low-frequency trends, respectively. However, the eigendecompositions, taken individually, do not necessarily lead to eigenvectors that combine both virtues. 

\subsection{Guiding Spectral Embedding Using a New Criterion}
\label{sssec:newcrit}

We hereby propose to further generalize the Slepian design in a number of ways. First, we relax the selection matrix $\ma S$ to a cooperation matrix $\ma M$ with diagonal elements that can take any non-negative real values $m_{l}\ge 0$, $l=1,\ldots,N$. This allows to gradually change the impact of a node on the analysis, between an enhanced ($m_l>1$), an unmodified ($m_l=1$) and a reduced ($m_l<1$) importance with respect to the selection matrix case.
Second, we combine the criteria of both already existing Slepian designs by subtracting the modified embedded distance from the energy concentration:

\begin{equation}
  \label{eq:newcrit}
   \zeta = \mu - \xi = \frac{\vc g^\top \ma M \vc g-\vc g^\top \ma L^{1/2} \ma M \ma L^{1/2} \vc g}{\vc g^\top \vc g} \quad\text{s.t. }\vc g \in \mathcal{B}_W.
\end{equation}
Third, we remove the bandlimit constraint and allow $\vc g$ to be any graph signal, which is an operational choice due to the joint optimization of both criteria, as will be illustrated and discussed later on.

Using the Taylor series approximation of the square root function, we derive $\ma L^{1/2}$ in terms of the adjacency matrix $\ma A$:
\begin{eqnarray}
\label{eq:L12}
  \ma L^{1/2}  = (\ma I-\ma A)^{1/2} &=& \ma I - \frac{1}{2} \ma A - \frac{1}{8} \ma A^2 - \frac{1}{16} \ma A^3 - \ldots \\
  &=& \ma I - \sum_{k=1}^{\infty} c_k \ma A^k,
\end{eqnarray}
with $c_k= \frac{(2k)!}{2^{2k}(k!)^2(2k-1)}.$ Details on the series expansion are discussed in Section~\ref{sssec:Taylor}. We can then further rewrite the internal part of the criterion (\ref{eq:newcrit}) as
\begin{eqnarray}
\label{eq:newcritexp}
  \ma M - (\ma I-\ma A)^{1/2}\ma M(\ma I-\ma A)^{1/2} & = & \sum_{k=1}^\infty c_k \left(\ma M\ma A^k+\ma A^k\ma M\right) \nonumber\\
  & & - \sum_{k_1=1}^\infty \sum_{k_2=1}^\infty \left(c_{k_1}c_{k_2}\right) \ma A^{k_1} \ma M \ma A^{k_2}.
\end{eqnarray}

By convention, the associated eigenvalues are sorted in decreasing order. Based on the fact that eigenvalues of the symmetric normalized Laplacian are greater or equal to 0 and lower or equal to 2, one can derive
${m_{\text{max}}\ge\zeta_1\ge \zeta_2 \ge \ldots \ge -2m_{\text{max}}}$, 
where $m_{\text{max}}$ is the highest cooperation value appearing in $\ma M$, using bounds from Corollary 2.4 in ~\cite{Lu2000eigvalbounds}.

In what follows, we will be considering the linear and quadratic approximations of the new criterion's eigenvalues: 
\begin{eqnarray}
  \zeta_{\text{lin}} & = & \frac{\vc g^\top \left(\frac{\ma M\ma A+\ma A\ma M}{2}\right) \vc g}{\vc g^\top\vc g} \\
  \zeta_{\text{quad}} & = & \frac{ \vc g^\top \left( \frac{\ma M\ma A+\ma A\ma M}{2} +  \frac{\ma M\ma A^2+\ma A^2\ma M}{8} - \frac{\ma A\ma M\ma A}{4} \right)\vc g}{\vc g^\top\vc g}.
\end{eqnarray}

Interestingly, the combination of both existing Slepian criteria leads to the emergence of the adjacency matrix $\ma A$ as the key player in our new formalism. In fact, when the cooperation matrix is the identity matrix, the criterion reverts to the eigendecomposition of $\ma A$ itself. 

\begin{figure}
\centering
\includegraphics[width=0.95\textwidth]{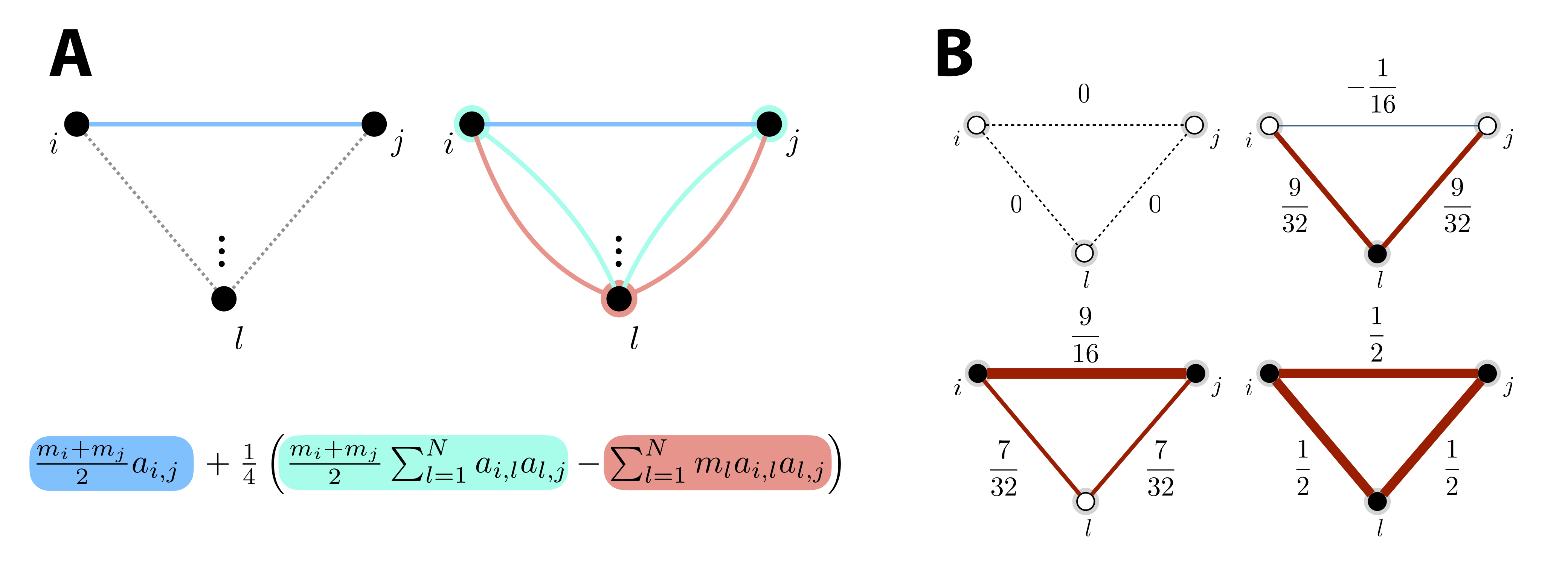}
\caption{\label{fig:modified-path-2} \textbf{A}. In the case of two nodes $i$ and $j$, the average of their cooperation weights yields the multiplying factor for $a_{i,j}$ (blue term). When a third node $l$ is added, the difference between average cooperation weight between nodes $i$ and $j$ (light blue term), and the cooperation weight of node $l$ (salmon term), multiplies the length-2 path and then also contributes to the output entry. \textbf{B}. In an example three-node network, output entries for different examples where cooperation weights are either set to $0$ (white nodes) or to $1$ (black nodes). Edge thickness is proportional to the output entry weight. Red strokes denote positive edge values, while blue strokes highlight negative edge values. All non-zero entries of the normalized adjacency matrix of the example network equal 1/2.}
\end{figure}

Let us now interpret the impact of the cooperation weights: obviously, an element $a_{i,j}$ of the adjacency matrix contains the weight of a direct path from $i$ to $j$. The linear approximation $\zeta_\text{lin}$ reweights such a direct path with the average $(m_i+m_j)/2$ of the cooperation weights that are attributed to nodes $i$ and $j$, as illustrated in Fig.~\ref{fig:modified-path-2}A (left half). Notice that paths where only one node has a cooperation weight equal to $0$ are still possible, as the other cooperation weight is then simply divided by two. 

As for the quadratic approximation, it takes into account length-2 paths between nodes $i$ and $j$. For instance, the sum of all length-2 paths between $i$ and $j$ can be read out from the squared adjacency matrix:

$$
  [\ma A^2]_{i,j} = \sum_{l=1}^N a_{i,l} a_{l,j} = \left< \mathbf{a}_{i,\cdot}, \mathbf{a}_{\cdot,j} \right>,
$$
where the inner product reveals the kernel interpretation of the length-2 walk matrix. Therefore, as illustrated in Fig.~\ref{fig:modified-path-2}A (right half), the term

$$
  [ \ma M\ma A^2+\ma A^2\ma M ]_{i,j} = (m_i + m_j) \sum_{l=1}^N a_{i,l} a_{l,j}
$$
reweights all length-2 paths by the summed cooperation weight between the start and end nodes, while subtracting the term

$$ 
   [ \ma A \ma M \ma A ]_{i,j} = \sum_{l=1}^N m_l a_{i,l} a_{l,j}
$$
penalizes the path according to the cooperation weight of node $l$ through which it passes.

Analogously, the term $\ma A^k$ in the criterion introduces modifications of $k$-length paths in the graph. However, for $k>N$, reweighting reduces to modifications of lower-length paths. The Cayley-Hamilton theorem implies that for every matrix $\ma A$ of size $N\times N$, the matrix $\ma A^N$ can be written as a linear combination of matrices $\ma A^k$ for $k=0,1,\ldots N-1$. By induction, it holds that $\ma A^k$ for every $k>N$ can also be written as a linear combination of the same set of $N$ matrices. Hence, modifications of paths longer than $N-1$ can be seen as a linear combination of additional modifications of paths of length 0 to $N-1$.

\section{Mathematical Considerations}
This section provides mathematical foundations supporting the methods and the results presented in this work. We start by discussing the link between the selection matrix and the eigenspectrum associated to the energy concentration criterion, and the relationship with the modified embedded distance criterion, using full bandwidth. Then, we provide a formal justification of the Taylor series approximation of the square root matrix function used in Eq. (\ref{eq:L12}), and discuss the error associated to this approximation.
\subsection{Eigenspectrum associated to the energy concentration criterion}
\label{sssec:mu}


For full bandwidth, the concentration matrix is defined as $\ma C=\ma U^\top\ma S\ma U$, where $\ma U$ is the matrix whose columns are eigenvectors of the graph Laplacian, and $\ma S$ is a diagonal selection matrix. Hence, the eigendecomposition of $\ma C$ is trivial: its eigenvectors are the rows of $\ma U$, and the eigenvalues of $\ma C$ correspond to the diagonal entries of $\ma S$, as can be seen from Fig. \ref{fig:criteria}A for $W=279$. 

\subsection{Eigenspectrum associated to the modified embedded distance criterion}
\label{sssec:xi}

We show that for full bandwidth, the number of zero eigenvalues of the modified embedded distance matrix, denoted $z_{\lambda}$, is lower-bounded by the number of zeros on the diagonal of the selection matrix, denoted $z_S$. To see this, consider the following decomposition of the modified embedded distance matrix $\ma C_{emb}$:

$$\ma C_{emb}=\ma L^{1/2}\ma S\ma L^{1/2}=\sum_{k=1}^{N-z_S}s_{i_k}l_{i_k}l_{i_k}^\top,$$
where $i_k$ is the index of the $k^{\text{th}}$ non-zero entry of the selection matrix $\ma S$, and $l_{i_k}$ denotes the $i_k$\textsuperscript{th} column vector of the matrix $\ma L^{1/2}$. From this expression, it can be seen that the rank of $\ma C_{emb}$ is at most $N-z_S$ and hence, $z_{\lambda}\geq z_S$. Equality holds when the set of vectors $\{l_{i_k}\}$ corresponding to the non-zero entries of $\ma S$ are linearly independent. This is the case for connected graphs, as any subset (with cardinality strictly less than $N$) of the columns of $\ma L^{1/2}$ is linearly independent. This relationship is observed in Fig. \ref{fig:criteria}B for $W=279$.

\subsection{Taylor series of matrix-valued functions}
\label{sssec:Taylor}

The Taylor expansion of $\ma L^{1/2}$ proposed in Eq. (\ref{eq:L12}) is derived using the scalar Taylor series of $f(x)=\sqrt{x}$ evaluated around the point $a=1$:
$$\sqrt{x}=1+\sum_{k=1}^{\infty}t_k(x-1)^k,$$
where $t_k= \frac{(-1)^{k-1}(2k)!}{2^{2k}(k!)^2(2k-1)}$ and $x\in \mathbb{R}, x>0$. The square root matrix of $\ma L$ then writes:
%
\begin{eqnarray}
\ma L^{1/2}&=&\ma U_L\ma \Lambda^{1/2}_L\ma U_L^\top\nonumber\\
&=&\ma U_L\begin{bmatrix}
1+\sum_{k=1}^{\infty}t_k(\lambda _1-1)^k &  &  \\
& \ddots & \\
&  & 1+\sum_{k=1}^{\infty}t_k(\lambda _N-1)^k \\
\end{bmatrix}\ma U_L^\top\nonumber\\
&=&\ma U_L(\ma I+\sum_{k=1}^{\infty}t_k(\ma \Lambda _L-\ma I)^k)\ma U_L^\top\nonumber.
\end{eqnarray}

Since the Laplacian and adjacency matrices are normalized, their eigenvalues verify $\ma \Lambda _L=\ma I-\ma \Lambda _A$ and their eigenvectors are equal ($\ma U_L=\ma U_A$) when ordered following increasing and decreasing eigenvalues, respectively. The previous equation finally reduces to:
\begin{eqnarray*}
\ma L^{1/2} & = & \ma I +\ma U_A(\sum_{k=1}^{\infty}t_k(-\ma \Lambda _A)^k)\ma U_A^\top \\
 & = & \ma I +\sum_{k=1}^{\infty}(-1)^kt_k\ma U_A\ma \Lambda _A^k\ma U_A^\top \\
 & = & \ma I -\sum_{k=1}^{\infty}c_k\ma A^k,
\end{eqnarray*}
\noindent where $c_k= \frac{(2k)!}{2^{2k}(k!)^2(2k-1)}$, which is the expression used in Eq. \eqref{eq:L12}.

Truncation of the Taylor series of a function $f(x)$ to a finite upper bound on $k\le K$ leads to an approximation error which can be estimated by the Lagrange form of the remainder

$$R_K(x)=\frac{f^{(K+1)}(y )}{(K+1)!}(x-1)^{K+1},$$
where the $(K+1)^{\text{th}}$ derivative is evaluated at the point $y$ found between $x$ and $1$. On the other hand, since the eigenvectors forming $\ma U_L$ are unit-norm vectors, the distance $d_K$ between a finite sum approximation of $\ma L^{1/2}$ and the true square root of the matrix is bounded as:

$$d_K=||\ma L^{1/2}-(\ma I -\sum_{k=1}^{K}c_k\ma A^k)||_F\le \sum_{i=1}^{N}|R_K(\lambda _i)|,$$
where $||\cdot||_F$ denotes the Frobenius norm. In the case of a first order Taylor approximation ($K=1$), we get:

$$d_1\le \sum_{i=1}^{N}\frac{|f^{(2)}(y _i)|}{2!}(\lambda _i-1)^2.$$


The eigenvalues $\lambda _i$ range from $0$ to $2$, and all contribute to the total approximation error $d_1$, with eigenvalues further from $1$ contributing more. Since the second-order derivative of the square root function increases as its argument approaches $0$, the most contributing factors of the error derive from Taylor approximation terms with near-zero eigenvalues. Hence, graphs whose Laplacian spectrum exhibits higher eigengaps in the lower band tend to have lower approximation error.

Finally, the Frobenius distance $d_{K,M}$ between the true proposed criterion $\ma M -\ma L^{1/2}\ma M\ma L^{1/2}$ and its approximation using a $K$\textsuperscript{th}-order Taylor approximation of $\ma L^{1/2}$ verifies:
$$d_{K,M}\le d_K ||\ma M||_F d_K,$$
where $||\ma M||_F$ corresponds to the Frobenius norm of the cooperation matrix. Hence, the upper bound on $d_{K,M}$ reduces as the nodes are given less importance; \textit{i.e.}, when the cooperation values get closer to 0.

\section{Results}

The \textit{C.~elegans} worm is an intensely studied model organism in biology. In particular, the wiring diagram of its 302 neurons has been carefully mapped during a long and effortful study~\citep{White1986}. Here, we use the graph that summarizes data from 279 somatic neurons (unconnected and pharyngeal neurons were excluded from the full diagram of 302 neurons), and combined connectivity from chemical synapses and gap junctions~\citep{Chen.2006}. The binary adjacency matrix $\ma A_{bin}$ with edge weights 0 or 1 has been symmetrically normalized with the degree matrix $\ma D$ into $\ma A=\ma D^{-1/2}\ma A_{bin}\ma D^{-1/2}$, as described in Section \ref{sec:methods_essential}. We retrieved the type of each neuron (sensory neuron, interneuron or motoneuron) from the WormAtlas database ({http://www.wormatlas.org/}). 

In their modeling work, \cite{Varshney2011} studied network properties of the worm connectome using different approaches, including Laplacian embedding. In particular, the topological view generated by mapping nodes on the first two eigenvectors with smallest non-zero eigenvalues already reveals interesting network organization (see Fig.~\ref{fig:principle}). The horizontal dimension ($\vc u_2$) 
mainly distinguishes the motoneurons from the head (right green circles) and from the ventral cord (left green circles). The vertical dimension ($\vc u_3$) reflects information flow from sensory neurons and interneurons of the animal's head (top) to the nerve ring and ventral cord circuitries (bottom).

\subsection{Eigenvalues of Different Criteria}

\begin{figure}
\centering
\includegraphics[width=\textwidth]{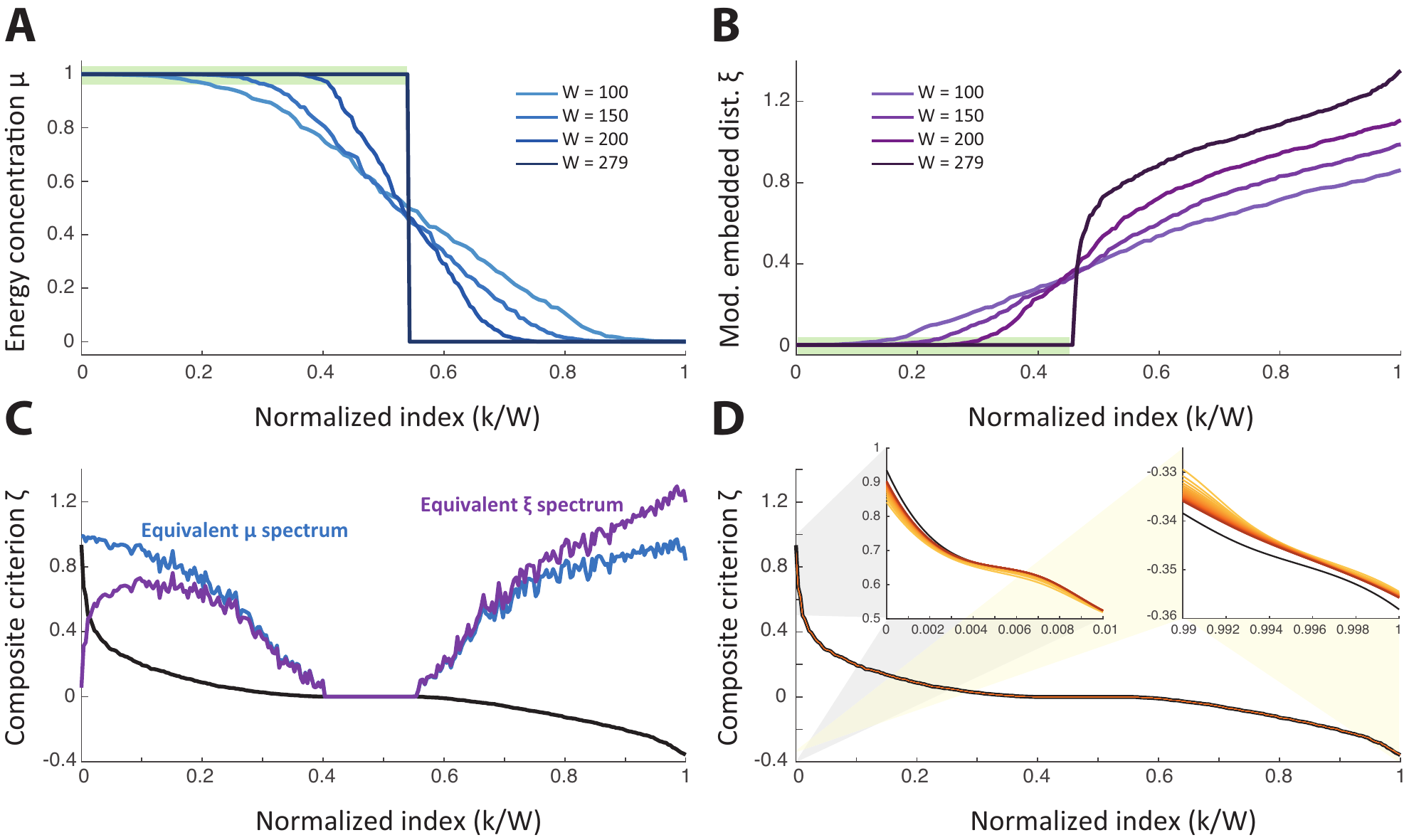}
\caption{\label{fig:criteria} Plots of eigenvalues obtained using different Slepian criteria: (\textbf{A}) energy concentration $\mu$, (\textbf{B}) modified embedded distance $\xi$, and (\textbf{C}) our new proposed criterion $\zeta$. For the first two cases, in which the design depends on a bandwidth parameter, eigenvalue spectra are plotted for $W=100$, $150$, $200$ and $279$ with increasingly lighter blue or purple shades, respectively. In the full bandwidth case, the shaded green areas highlight eigenvalues linked to optimal solutions of the respective criteria (see Sections~\ref{sssec:mu} and~\ref{sssec:xi} for the associated mathematical derivations). In the third case, equivalent $\mu$ and $\xi$ eigenspectra are plotted in blue and purple on top of the $\zeta$ one. The full $\zeta$ eigenspectrum is also compared to approximations obtained through Taylor series of increasing order (\textbf{D}), from linear to order 20, as depicted by increasingly darker brown curves. The two smaller plots are insets sampled at the start and at the end of the main plot, respectively.}
\end{figure}

To illustrate the eigenvalues obtained with the existing Slepian designs, as well as the newly proposed criterion, we considered the 128 motoneurons and ``unselected'' them by setting their respective entries in $\ma S$ to $0$. We applied the original, concentration-based Slepian design for different bandwidths $W=100, 150, 200, 279$, the latter corresponding to full bandwidth. The eigenvalues $\mu_k$, which reflect energy concentration in the 151 remaining neurons, are shown in Fig.~\ref{fig:criteria}A. The characteristic behavior of classical Slepians is preserved for the graph variant; \textit{i.e.}, eigenvalues cluster around $1$ and $0$ for well and poorly concentrated eigenvectors, respectively, and the phase transition occurs more abruptly at higher bandwidth. For full bandwidth, perfect concentration becomes possible, and the problem degenerates in retrieving two linear subspaces of 151 and 128 dimensions spanned by eigenvectors with concentration $1$ and $0$, respectively (see Section~\ref{sssec:mu} for a proof on the number of distinct eigenvalues). In practical terms, for high but not full bandwidth, the ``interesting'' eigenvectors with large concentration correspond to the part indicated by the green area on the plot, and become numerically indistinguishable. A few indicative examples of Slepian vectors across bandwidths are displayed in Supplementary Fig.~\ref{fig:sf1}C.

Next, we applied the modified Slepian design inspired by the Laplacian embedded distance. As shown in Fig.~\ref{fig:criteria}B, the eigenvalues $\xi_k$ reflect the modified embedded distance, which we now want to minimize. For increasing bandwidth (darker curves), its smallest values can be made lower; however, the subset of nodes with $S_{i,i}$ entries set to $0$ is also described by eigenvectors with small eigenvalues. This becomes even clearer at full bandwidth, a case for which a subspace of 128 dimensions spanned by eigenvectors with a modified embedded distance of $0$ is retrieved, as indicated by the green area in Fig.~\ref{fig:criteria}B and explicitly demonstrated in Section~\ref{sssec:xi}. Some examples of Slepians across bandwidths can be seen in Supplementary Fig.~\ref{fig:sf1}D.

The degeneracies of the Slepian designs at full bandwidth are instructive about the opposing effects of maximizing energy concentration and minimizing modified embedded distance; \textit{i.e.}, the subspaces indicated by the green areas in Figs.~\ref{fig:criteria}A and B, which are optimal for the corresponding criteria, are actually different ones, representing signals on sensory and interneurons (151 nodes) on the one hand, and on motoneurons (128 nodes) on the other hand (compare Supplementary Figs.~\ref{fig:sf1}C and D, first rows). This leads us to the eigenvalues $\zeta_k$ of the proposed criterion, as shown in Fig.~\ref{fig:criteria}C (black curve). 

The maximum eigenvalue peaks close to $1$, a case reflecting jointly high equivalent $\mu_k$ (blue curve) and low equivalent $\xi_k$ (purple curve); \textit{i.e.}, a high energy concentration at the same time as a low modified embedded distance (low localized graph frequency) within $\mathcal{S}$. The low amount of such solutions shows that it is difficult to conceal high energy concentration and small modified embedded distance.

As values of $\zeta_k$ decrease, we first observe a rise in modified embedded distance (eigenvectors remain reasonably concentrated within $\mathcal{S}$, but rapidly exhibit a larger localized graph frequency), and then a decrease of both $\mu_k$ and $\xi_k$, which indicates that eigenvectors become less concentrated within the subset of interest. Afterwards, we observe a regime in which both quantities are null at the same time; that is, a subspace spanned by eigenvectors that are fully concentrated outside $\mathcal{S}$. Notice that this set of eigenvectors is now ``pushed away'' from the meaningful low $\xi_k$ ones, and lie in the middle of the spectrum. Finally, the sign of $\zeta_k$ switches, and the right hand side of Fig.~\ref{fig:criteria}C denotes eigenvectors of increasing concentration within $\mathcal{S}$ and localized graph frequency, the latter effect dominating over the former.

Interestingly, computing the eigenspectrum using a linear approximation of the criterion matrix (Fig.~\ref{fig:criteria}D, light brown curve) leads to very similar results, which only slightly vary for the largest eigenvalues. When the approximation order is increased up to 20 (increasingly dark brown curves), this low error further diminishes, although a mild difference remains with the ground truth. Inspection of the Slepian vectors related to several locations of the eigenspectrum (Supplementary Fig.~\ref{fig:sf2}) confirmed that the only salient differences actually involved the first Slepian vector (largest eigenvalue one).

\subsection{Topology Revealed by Guided Spectral Analysis}

We now guide the spectral analysis to focus on the three different types of neurons. For instance, when focussing on the role of the sensory neurons, we gradually decrease the cooperation weights of interneurons and motoneurons from $1$ to $0$. For each setting, we then visualize the topology revealed by the guided analysis by projecting the nodes on the eigenvectors with the second and third largest eigenvalues. We build the trajectory of each node through this two-dimensional embedding, after applying the Procrustes transform~\citep{Procrustes} to compensate for any irrelevant global transformations. As a complementary visualization, note that we provide the start, intermediate and end points of each trajectory as separate figures in Supplementary Fig.~\ref{fig:sf3}. Finally, k-means clustering was performed on the nodes in focus at the end point embedding of trajectories, producing sets of clusters given in Supplementary Fig.~\ref{fig:clusters} and Supplementary Tables \ref{tab:sensory}-\ref{tab:motor} (see Section~\ref{subsec::clustering} for details). Example visualizations when resorting to different Slepian vectors are provided in Supplementary Fig.~\ref{fig:sf4}.

In Figs.~\ref{fig:sensory}A and B, the trajectories are depicted when focussing on the sensory neurons by attributing cooperation weights to the other types of neurons ranging from $1$ to $0.5$, and from $0.5$ to $0$, respectively. During the first half (Fig.~\ref{fig:sensory}A), the network organization is only slightly altered with respect to the initial view of Fig.~\ref{fig:principle}; \textit{i.e.}, the sensory neurons move slightly more to the periphery, while the interneurons and motoneurons move to the origin. In the second part of the trajectory (Fig.~\ref{fig:sensory}B), a major split occurs in the bottom right branch of Fig.~\ref{fig:sensory}A between the left and right versions of a whole series of neurons, while the bottom left branch neurons move back to the center of the coordinate frame. The cell types found in the top branch are amphid neurons, whereas the rest of the sensory neurons split into their left and right counterparts located in the left and right bottom branches. The clusters found by the k-means approach (see Supplementary Table~\ref{tab:sensory}) include a group of $5$ bilateral amphid neurons (AWA, AWC, ASE, ASI and AFD; cluster C\textsubscript{3}) and $6$ other clusters, 2 of which span the bottom left and right sub-branches (clusters C\textsubscript{5} and C\textsubscript{2}).


As described in Section \ref{sssec:newcrit}, since paths through nodes with cooperation weights set to 0 are still considered by the proposed criterion, the embedding focusing on a particular subtype of neurons can still include functionally distinct cells as clearly standing out in the visualization. For instance, in addition to the above clustering of sensory neurons in Fig.~\ref{fig:sensory}B, we notice the segregation of the bilateral RIP interneurons towards the left and the right branch. This shows that the embedding does not neglect nodes outside the focus, even when their cooperation weight is set to 0.

\begin{figure}
\centering
\includegraphics[width=1\textwidth]{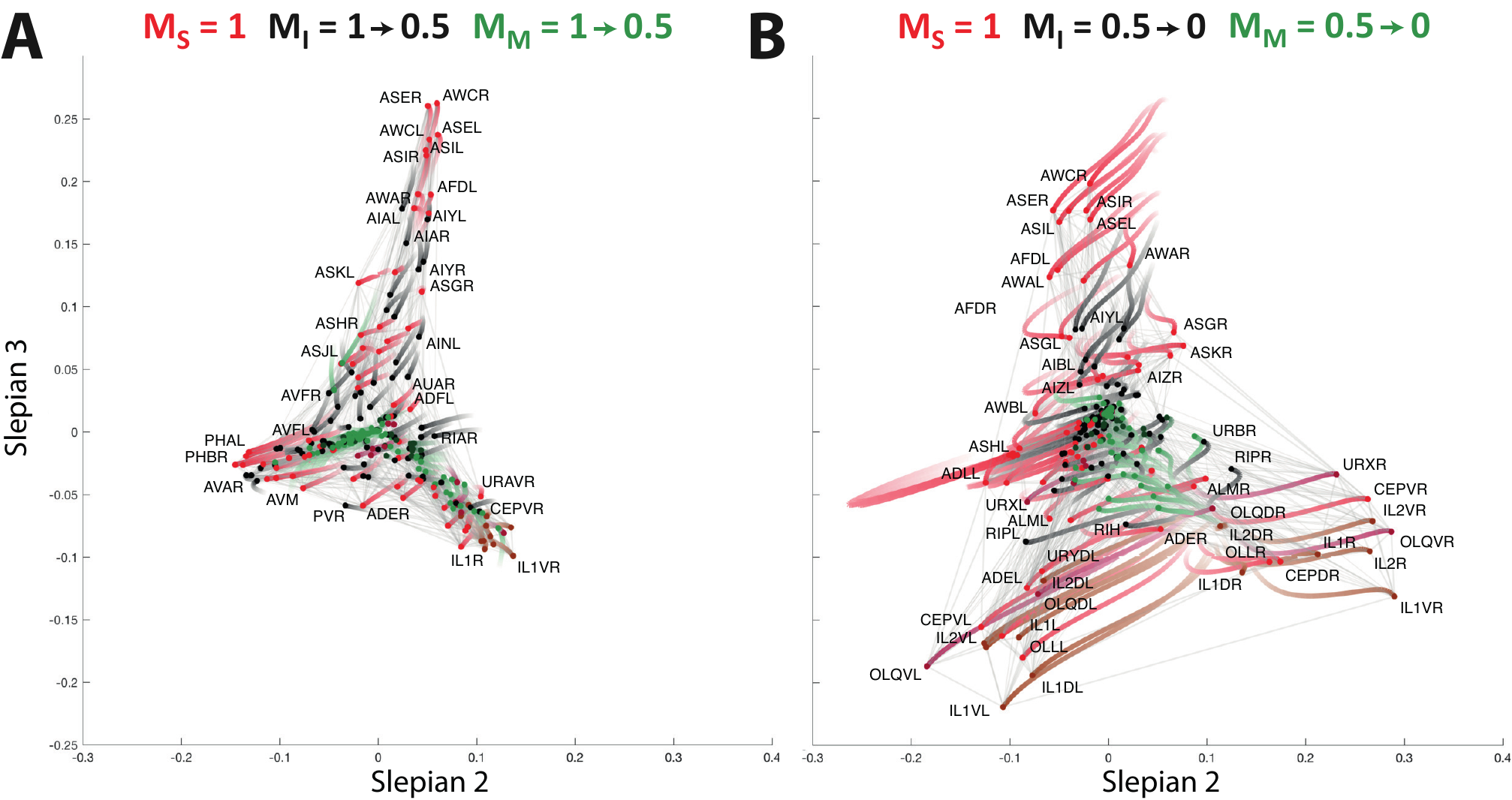}
\caption{\label{fig:sensory} Focussing on the sensory neurons by reducing the cooperation weights of the interneurons and motoneurons (\textbf{A})~from $1$ to $0.5$, and (\textbf{B})~from $0.5$ to $0$. The trajectory of a neuron is represented by a colour change from light to dark tones, and dots represent final positions. Note that the starting configuration in (\textbf{A}) is identical to the representation in Fig.~\ref{fig:principle}. Cells are labeled according to \cite{Varshney2011}.}
\end{figure}

\begin{figure}
\centering
\includegraphics[width=1\textwidth]{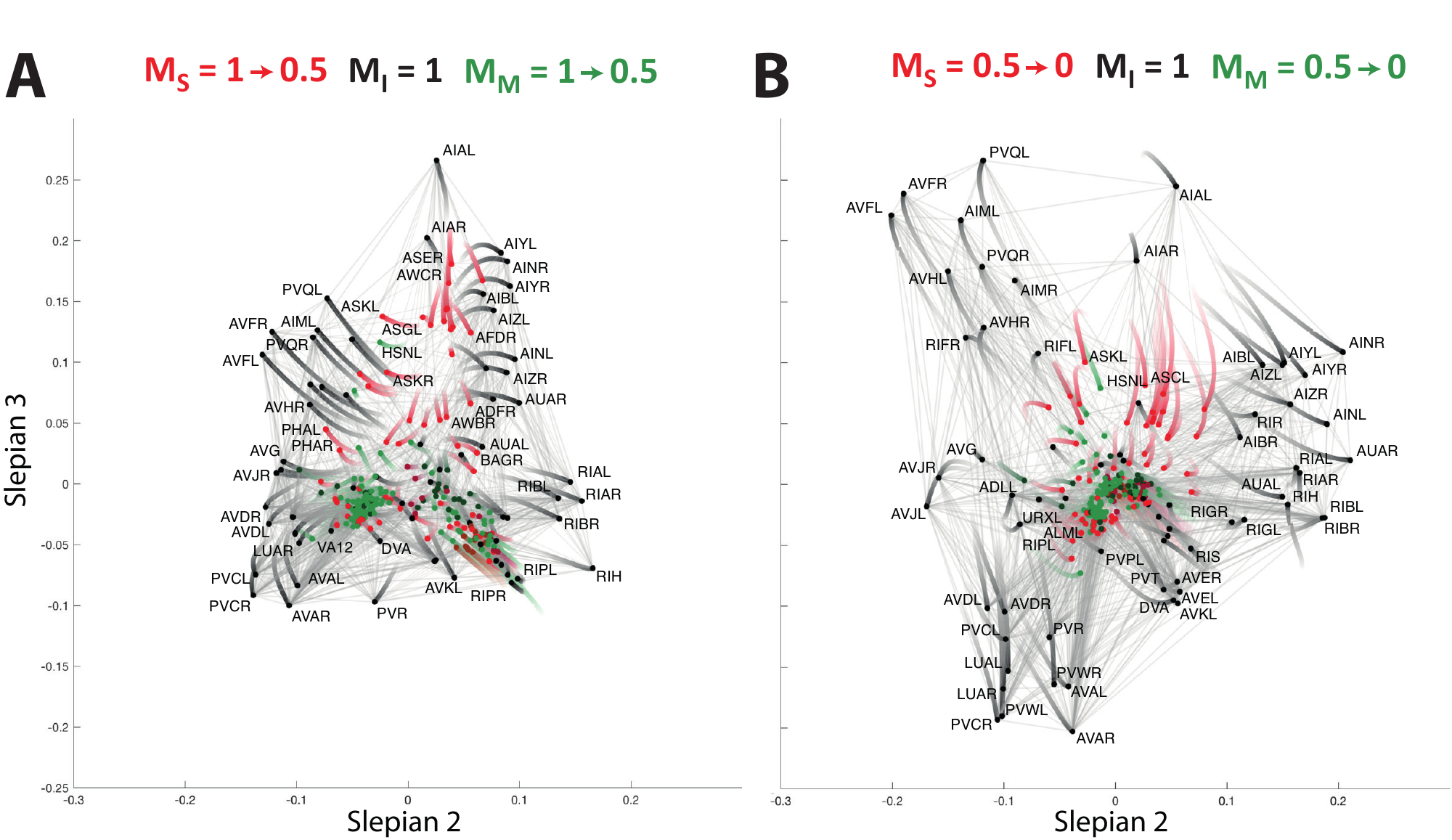}
\caption{\label{fig:inter} Focussing on the interneurons by reducing the cooperation weights of the sensory neurons and motoneurons (\textbf{A})~from $1$ to $0.5$, and (\textbf{B})~from $0.5$ to $0$. The trajectory of a neuron is represented by a colour change from light to dark tones, and dots represent final positions. Note that the starting configuration in (\textbf{A}) is identical to the representation in Fig.~\ref{fig:principle}. Cells are labeled according to \cite{Varshney2011}.}
\end{figure}

In Figs.~\ref{fig:inter}A and B, we then focus on the interneurons by reducing the cooperation weights of sensory neurons and motoneurons in two steps. As expected, the interneurons move towards the periphery. Their organization does not seem to be dominated by left \textit{versus} right variants, as we found for sensory neurons, but rather by a set of well-defined clusters related to their functional involvement in the \textit{C. elegans} neuronal circuitry (see Supplementary Table~\ref{tab:inter}): in the first quadrant, we find the isolated AIA bilateral pair (cluster C\textsubscript{4}). Moving clockwise, a larger cluster of neurons includes the bilateral AIY, AIZ, AIN, AIB, RIA, RIB, AUA and the single neurons RIR and RIH (cluster C\textsubscript{3}). Next we find a cluster including AVE, AVK, RIG, PVT, DVA and other neurons located closer to the origin of Fig.~\ref{fig:inter} (cluster C\textsubscript{5}), before reaching another large ensemble of neurons including the bilateral AVA, AVD, LUA, PVC, PVW, and the single neuron PVR (cluster C\textsubscript{6}). Moving back upwards, cluster C\textsubscript{1} contains the bilateral AVB, AVJ, BDU, the single neuron AVG, and PVPR, whose left counterpart PVPL belongs to cluster C\textsubscript{5}, thus standing as the only bilateral pair of neurons split into different clusters. Finally, we reach the last group of cells containing the bilateral RIF, AVH, AIM, PVQ and AVF (cluster C\textsubscript{2}).


\begin{figure}
\centering
\includegraphics[width=1\textwidth]{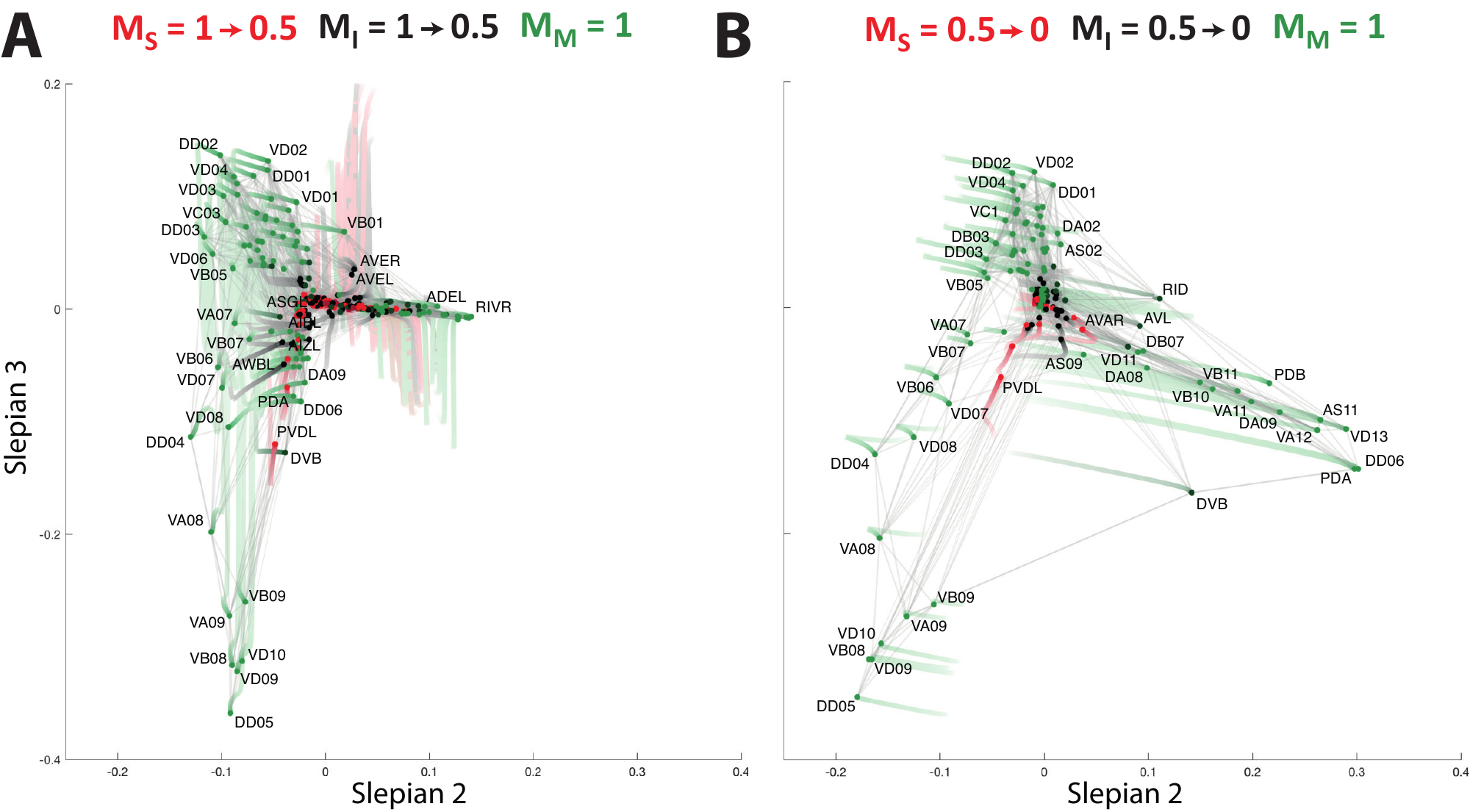}
\caption{\label{fig:motor} Focussing on the motoneurons by reducing the cooperation weights of the sensory neurons and interneurons (\textbf{A})~from $1$ to $0.5$, and (\textbf{B})~from $0.5$ to $0$. The trajectory of a neuron is represented by a colour change from light to dark tones, and dots represent final positions. Note that the starting configuration in (\textbf{A}) is identical to the representation in Fig.~\ref{fig:principle}. Cells are labeled according to \cite{Varshney2011}.}
\end{figure}

Finally, in Figs.~\ref{fig:motor}A and B, the organization of motoneurons is examined. Already in the first step (Fig.~\ref{fig:motor}A), when reducing the cooperation weights of the sensory and interneurons from $1.0$ to $0.5$, we observe much stronger changes than in the previous cases. In particular, the initial organization completely collapses and the left branch of the motoneurons spreads out. This branch then develops into a peripheral organization when further decreasing the cooperation weights (Fig.~\ref{fig:motor}B), with three main subsets of neurons and ambiguous positioning of the cell DVB between the left and the right bottom branches. K-means clustering into optimal cell groups captured this architecture into $7$ smaller clusters (Supplementary Table \ref{tab:motor}): clusters C\textsubscript{4} and C\textsubscript{7} spanned top neurons, clusters C\textsubscript{2} and C\textsubscript{3} included the bottom left branch neurons, and clusters C\textsubscript{5} and C\textsubscript{6} contained the bottom right branch cells.


\section{Discussion}

\subsection{Beyond Original Slepian Designs}

The originality of our approach lies in providing a new and simple way to guide graph spectral analysis. Inspired by graph Slepians, we propose a novel criterion that combines energy concentration and modified embedded distance, taking into account cooperation weights that can gradually increase or decrease the importance of selected nodes. The new criterion lets the adjacency matrix emerge as the central graph operator, instead of the Laplacian, and is operational at full bandwidth.

This is surprising at first sight, because neither of the conventional Slepian criteria is practical without the bandlimit constraint. For the energy concentration with binary cooperation weights, as shown in Fig.~\ref{fig:criteria}A for an illustrative example on the \textit{C.~elegans} connectome, full bandwidth leads to two eigenvalues ($1$ and $0$), the dimensionality of the corresponding subspaces being the number of nodes with cooperation weight $1$ and $0$, respectively. For the modified embedded distance, as shown in Fig.~\ref{fig:criteria}B, full bandwidth creates a subspace with eigenvalue $0$ of dimensionality equal to the number of nodes with cooperation weight $0$. Therefore, subtracting both criteria leads to opposing objectives; \textit{i.e.}, at full bandwidth, an energy concentration of $1$ encodes the subspace for nodes with weight $1$, while a modified embedded distance of $0$ encodes the subspace for nodes with weight $0$. 

The obtained eigenspectrum for the new criterion, shown in Fig.~\ref{fig:criteria}C, illustrates that only a few eigenvectors are able to combine high energy concentration with low modified embedded distance, a counterbalance that can be further revealed by measuring $\mu$ and $\xi$ separately for these new eigenvectors. Such a large eigengap is also good news for numerical computation of the leading eigenvectors for large graphs when relying upon efficient large-scale solvers~\citep{Lehoucq.1996} implemented in widely available software libraries such as ARPACK. 

Intriguingly, the approximation error was already low using a linear approximation, and did not noticeably decrease further, except for the first Slepian vector, when resorting to higher-order terms (see Fig.~\ref{fig:criteria} and Supplementary Fig.~\ref{fig:sf2}). Modifying the importance of a node \textit{via} the corresponding cooperation value affects all-length paths through that node according to the series expansion from Eq. (\ref{eq:newcritexp}), where the power of $\ma A$ in each term corresponds to the affected path length. Once we restrict the criterion to a linear approximation, the only paths whose importance is changed are those of length 1. This does not mean that other paths are not included in the graph analysis, but rather that they are included with their original (unmodified) effect on the topology. Low error of linear approximation suggests that the highest percentage of topological importance of a node falls into the importance of its length-1 paths. Further, a slightly higher error at eigenvectors with the highest $\zeta$ may be explained similarly: not modifying higher order paths produces greater error at these eigenvectors because of their increased relative importance due to the fact that high $\zeta$ eigenvectors tend to be very smooth (even approaching a constant signal); thus, in order to even out the values at all nodes in the process, one needs to "reach" far enough.

The proposed criterion should not be confused with the Sobolev norm that is sometimes used to regularize graph signals \citep{Mahadevan.2006}. Specifically, in the case of $\ma M=\ma I$, our criterion of Eq. (\ref{eq:newcrit}) applied to $\vc g$ reverts to $\ma g^\top\ma g - \ma g^\top\ma L\ma g$, whereas the Sobolev norm of $\vc g$ reads $\ma g^\top\ma g + \ma g^\top\ma L\ma g$. The difference in the sign of the second term introduces significantly distinct optimization goals regardless of the apparent similarity of the two expressions.

As for future extensions of our approach, one could envisage to dig into the relationship with graph uncertainty principles~\citep{Agaskar.2013,Tsitsvero.2016,Teke.2017}, to consider statistical resampling for graphs~\citep{pirondini1601}, or to focus on the discovery of hierarchical graph structure~\citep{Arenas.2008, Irion.2014} by gradual refinement of the subgraph. The design could also be extended to directed graphs using recent extensions of spectral decompositions in this context~\citep{Sandryhaila.2013,Mhaskar.2018}.

\subsection{Gaining Insights on \textit{C. elegans}}
\label{ssec:disc_celegans}

The application of our newly developed approach to the \emph{C. elegans} connectome enabled to confirm past findings from the literature, and to shed light on additional cellular targets and groupings that may deserve further experimental analyses. At the level of sensory neurons (Fig.~\ref{fig:sensory}, Supplementary Fig.~\ref{fig:sf3}A and Supplementary Fig.~\ref{fig:clusters}A), seven clusters were extracted, collectively accounting for the three branches evident in Fig.~\ref{fig:sensory}: the top branch made of twelve (including the thermosensor AFD) pairs of amphid neurons (at \textit{y}-coordinate greater than $0.04$), and other cells split into the left and right bottom branches. Interestingly, one of the clusters found by k-means included five pairs of bilateral amphid neurons:  AWA and AWC involved in odortaxis \citep{Bargmann1993,Li2012}, the thermosensor AFD \citep{Mori1995}, and ASE and ASI implicated in chemotaxis \citep{Bargmann1991,Luo2014}. These neurons act as low-order sensors, whose extraction as a separate cluster inside the amphid group may suggest new information worth further exploration.

\begin{figure}
\centering
\includegraphics[width=0.85\textwidth]{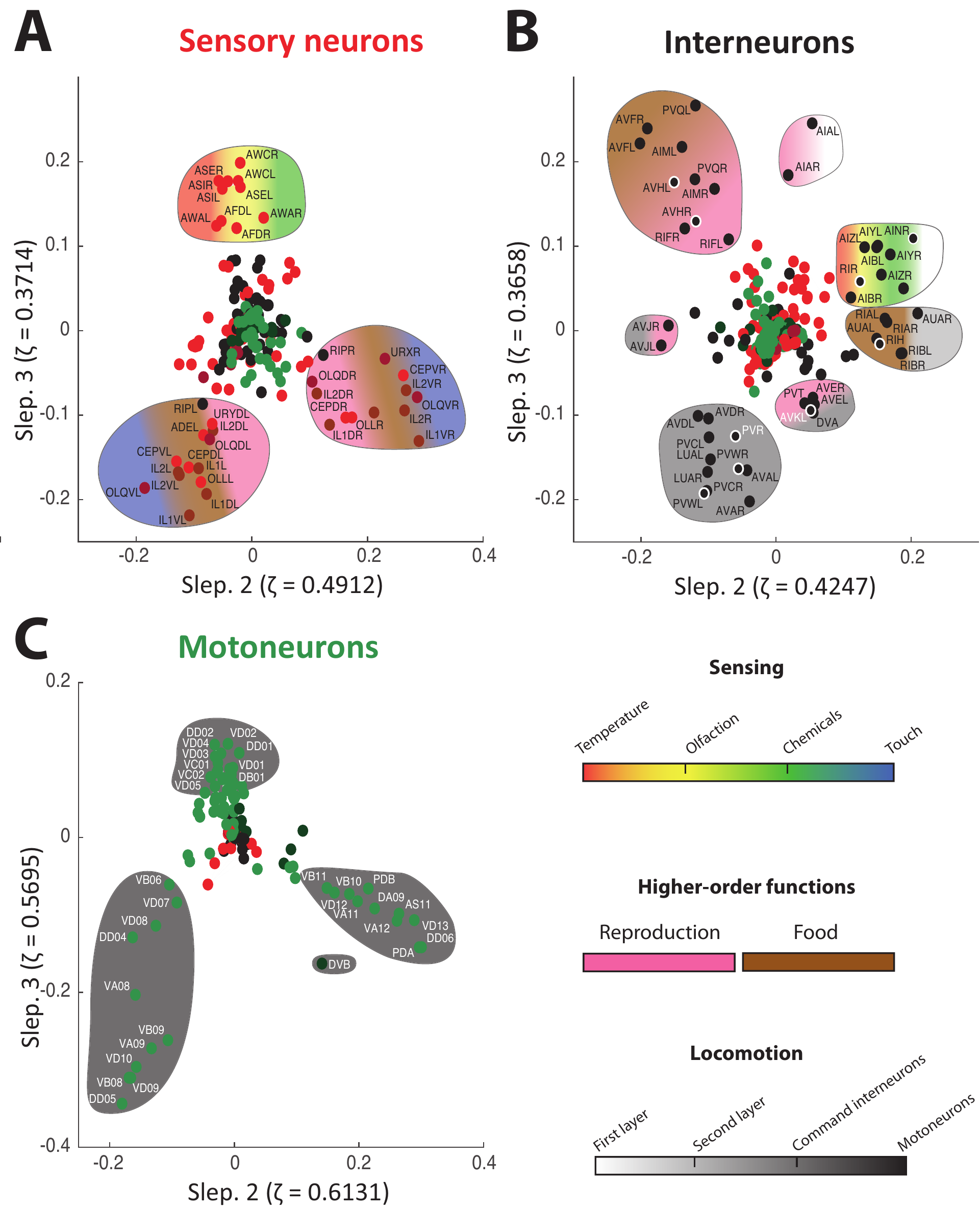}
\caption{\label{fig:summary} Summary of the main functions operated by the sensory neurons (\textbf{A}), interneurons (\textbf{B}) and motoneurons (\textbf{C}) unraveled by guided spectral analysis. Clusters of neurons discussed in Section~\ref{ssec:disc_celegans} are delineated and color coded according to their main roles: this may be in \textit{sensing} (thermosensation in red, olfactory sensation in yellow, chemosensation in green and mechanosensation in blue), \textit{higher-order functions} (reproduction in pink, food responses in brown), or \textit{locomotion} (from first cellular relays to effector motoneurons in increasingly darker shades of gray). A gradient in the color coding indicates that more than one function is performed by neurons from a given cluster. Neurons that could not be clearly related to the rest of the unraveled circuitry are encircled in white.}
\end{figure}

The lower branches in Fig.~\ref{fig:sensory} split the neurons into their right and left counterparts, thus extracting relevant somatic information. These neurons act as higher-order sensing apparatus as compared with amphid neurons: IL1 and OLQ have jointly been implicated in the worm foraging response \citep{Hart1995}; CEP and ADE are involved in the response upon food sensing \citep{Sawin2000}; URX, URY and OLL are linked to the reproductive drive \citep{Barrios2012}, and so on. The split between low and high order sensing is summarized in Fig. \ref{fig:summary}A.

Further inspection of the branches (Supplementary Fig.~\ref{fig:sf5}A) showed that the left-right segregation involved chemical synapses, but not gap junctions. Also, Supplementary Fig.~\ref{fig:sf4} (second row) shows that for higher-order Slepian vectors (fourth and fifth), additional contributors emerge, such as the bilateral PHA/PHB. This suggests that the approach finds different subgroups of higher-order sensory neurons depending on the choice of the embedding eigenvectors. The biological/functional intepretation of the exact clusters asks for a more detailed analysis of the subgroups of neurons. Finally, the emergence of RIP interneurons in the embedding (Fig.~\ref{fig:sensory}) points towards an important role of the sensory neurons yet to be explained, possibly in connection with their presynaptic inputs from IL1 \citep{White1986}.

Turning to interneurons (Fig.~\ref{fig:inter}, Supplementary Fig.~\ref{fig:sf3}B and Supplementary Fig.~\ref{fig:clusters}B), we notice a trend of grouping neurons at the same command-chain level. Starting from the top of Fig.~\ref{fig:inter}, we find AIA, AIB, AIY and AIZ jointly known for their role on locomotory behaviour and acting as a first-relay drives  \citep{Wakabayashi2004,Gray2005}. Moving clockwise, we find RIA and RIB acting as second-layer intermediates, and further on, neurons such as AVE, and in the next cluster AVAL and AVD, all being command interneurons \citep{Hobert2003,Haspel2010,Kawano2011}. The trend of following the locomotory pathway clockwise in the embedding space suggests that the approach targets relevant information about the neural system. However, the exact compact clusters in Supplementary Fig.~\ref{fig:clusters}B need further elaboration. Some of the interesting findings worth exploring would be the unexplained grouping of the scarcely studied RIR neuron \citep{Hobert2002} with the cluster of cells including AIB and AIY, or the grouping of PVR and LUA \citep{Chalfie1985,Wicks1995} with locomotion-regulating neurons such as AVD and AVA.


Considering motoneurons (Fig.~\ref{fig:motor}, Supplementary Fig.~\ref{fig:sf3}C and Supplementary Fig.~\ref{fig:clusters}C), the embedding positions fit somatic location (see Supplementary Fig.~\ref{fig:sf6}): a spiral beginning at the origin, turning right, then moving clockwise and ending in the top branch follows the postero-anterior direction. This confirms that the approach has extracted meaningful information. However, the exact split between the three branches as well as the k-means clustering into the seven ensembles remains unclear, since, from preliminary explorations, we find both A-type and B-type cholinergic motoneurons and the inhibitory D-type motoneurons in all clusters. Finally, DVB deserves further attention \citep{Schuske2004} due to its isolated location between the two bottom branches.

In Fig. \ref{fig:motor}B, two sensory nodes stick out the furthest away from the center; \textit{i.e.}, towards the lower left and right branches of motoneurons. These are PVD and PHC neurons, responsible for nociceptive mechano- and thermosensation, respectively. The locations of these nodes in the embedding may be linked to the fact that harmful nociceptive stimuli induce a locomotory response. As in the case of RIP interneurons emerging in the focused embedding of sensory neurons, we once again confirm the ability of the proposed approach to extract important nodes even when their cooperation weight was initially set to 0.



In summary, as illustrated in Fig.~\ref{fig:summary}, all three types of neurons found in the \textit{C. elegans} nematode could be arranged in a meaningful hierarchy thanks to the introduced guided graph spectral embedding. Sensory neurons were separated between first-order and higher-order sensors. Different levels of processing of motor functions were distinguished (see the gradient from white to dark gray tones going clockwise in Fig.~\ref{fig:summary}B), with the eventual recruitment of motoneurons, which have been separated on the basis of somatic location. Future analyses will allow the study of different types of neurons through more elaborate combinations of focused nodes. In addition, it will be interesting to see whether future experimental work can shed light on some of the neurons that were extracted here without being yet extensively documented in the literature, such as AVKL or RIR.

\subsection{Perspectives for Future Uses}
\label{ssec:future_uses}

The proposed graph embedding provides a simple, yet powerful approach to visualization and, if combined with clustering techniques, to the extraction of meaningful subgraphs from any graph-modeled dataset. In neuroimaging, focusing on a specific subgraph of interest (by setting the appropriate cooperation values) can direct research onto clinically relevant concepts, such as the medial temporal lobe and limbic structures for human brain imaging studies comparing healthy controls and Alzheimer patients~\citep{Krasuski1998}. Be it using the structural or the functional connectome for analyses~\citep{Contreras2015}, features such as cluster size and/or the inclusion of specific nodes (brain regions) in a cluster may become biomarkers for an early diagnosis or prediction of the disease.

Furthermore, graph modeling of the human brain is frequently employed to extract important nodes/brain regions and to identify their topological roles, such as a provincial/connector hubs suggesting clinically significant functional roles~\citep{vandenHeuvel2013}. Doing so requires the use of diverse node centrality measures, such as degree or betweenness centrality. On the other hand, entries of the proposed Slepian eigenvectors may be interpreted as higher-order spectral centrality measures relative to the focused subgraph, and for the special case $\ma M=\ma I$, the eigenvector corresponding to the highest positive eigenvalue reverts to the eigenvector centrality~\citep{Newman.2010}. Hence, if clustering of a dataset based on the proposed embedding coordinates reveals nodes distant from the rest of the graph, it is suggested that those nodes exhibit a hub-like role when the focused subgraph is considered more important than the rest of the graph. For example, the AIA pair in the discussed \textit{C. elegans} example emerges as a separate cluster in Fig. \ref{fig:inter} and Supplementary Fig. \ref{fig:clusters}B, where the focus is set on interneurons. Its role as a hub can be confirmed by the high number of connections to the set of amphid neurons, and a small number of connections to the other cells, as compared to the rest of the interneurons. Identification of hubs and/or peripheral nodes with respect to other similar type nodes may lead to a better understanding of the functional role of both neurons and brain regions, depending on the inspected dataset.

\acknowledgments
This work was supported in part by each of the followings: the CHIST-ERA IVAN project (20CH21\_174081), the Swiss National Science Foundation (project 200021\_175506), the Center for Biomedical Imaging (CIBM) of the Geneva - Lausanne Universities and the EPFL, and the Leenaards and Louis-Jeantet Foundations.

\authorcontributions 
DVDV designed the study, implemented the initial models, and ran the first analyses. MP derived the mathematical proofs, which were verified by RL. TB ran detailed analyses on the \textit{C.~elegans} connectome and contributed to their interpretation. DVDV wrote the initial draft of the manuscript. All authors read and revised the manuscript. 

\newpage

\section{Supportive Information}
\subsection{Results of k-means clustering}
\label{subsec::clustering}

In Fig. \ref{fig:clusters}, we present the proposed embedding from Figs. \ref{fig:sensory}-\ref{fig:motor} and clusters of nodes with cooperation weight $1$ derived by the k-means approach with 20 repetitions. Dimensionality of the considered data points was set to $2$, \textit{i.e.} entries of the two Slepian eigenvectors were used for clustering -- the second and the third. The Silhouette method~\citep{Rousseeuw1987} was used to estimate the optimal number of clusters as the one which produces the minimal number of negative silhouette values. Convex hulls of each found cluster are represented by dashed black lines. The exact lists of neurons assigned to each cluster, for the three investigated cell types, are provided in Supplementary Tables \ref{tab:sensory}, \ref{tab:inter} and \ref{tab:motor}.

\begin{table}[ht]
	\caption{\label{tab:sensory}Sensory neurons of the \textit{C. elegans}. Columns correspond to clusters derived by optimized k-means.}
	\scriptsize
	\centerline{\begin{tabular}{ccccccc} \text{C\textsubscript{1}} & \text{C\textsubscript{2}} & \text{C\textsubscript{3}} & \text{C\textsubscript{4}} & \text{C\textsubscript{5}} & \text{C\textsubscript{6}} & \text{C\textsubscript{7}} \\ \text{AVM} & \text{CEPDR} & \text{AFDL} & \text{PHAL} & \text{ADEL} & \text{ADFR} & \text{ADFL}\\ \text{PDEL} & \text{CEPVR} & \text{AFDR} & \text{PHAR} & \text{ADER} & \text{ADLR} & \text{ADLL}\\ \text{PDER} & \text{IL1DR} & \text{ASEL} & \text{PHBL} & \text{ALMR} & \text{ASGL} & \text{ALA}\\ \text{PHCL} & \text{IL1R} & \text{ASER} & \text{PHBR} & \text{CEPDL} & \text{ASGR} & \text{ALML}\\ \text{PHCR} & \text{IL1VR} & \text{ASIL} &   & \text{CEPVL} & \text{ASHR} & \text{ALNL}\\ \text{PLML} & \text{IL2DR} & \text{ASIR} &   & \text{IL1DL} & \text{ASKL} & \text{ALNR}\\ \text{PLMR} & \text{IL2R} & \text{AWAL} &   & \text{IL1L} & \text{ASKR} & \text{AQR}\\ \text{PVM} & \text{IL2VR} & \text{AWAR} &   & \text{IL1VL} & \text{AWBL} & \text{ASHL}\\   & \text{OLLR} & \text{AWCL} &   & \text{IL2DL} & \text{AWBR} & \text{ASJL}\\   & \text{OLQVR} & \text{AWCR} &   & \text{IL2L} &   & \text{ASJR}\\   & \text{URXR} &   &   & \text{IL2VL} &   & \text{BAGL}\\   &   &   &   & \text{OLLL} &   & \text{BAGR}\\   &   &   &   & \text{OLQDL} &   & \text{FLPL}\\   &   &   &   & \text{OLQDR} &   & \text{FLPR}\\   &   &   &   & \text{OLQVL} &   & \text{PLNL}\\   &   &   &   & \text{URYDL} &   & \text{PLNR}\\   &   &   &   & \text{URYDR} &   & \text{PQR}\\   &   &   &   & \text{URYVL} &   & \text{PVDL}\\   &   &   &   & \text{URYVR} &   & \text{PVDR}\\   &   &   &   &   &   & \text{SDQL}\\   &   &   &   &   &   & \text{SDQR}\\   &   &   &   &   &   & \text{URXL} \end{tabular}}
\end{table}

\begin{table}[ht]
	\caption{\label{tab:inter}Interneurons of the \textit{C. elegans}. Columns correspond to clusters derived by optimized k-means.}
	\scriptsize
	\centerline{\begin{tabular}{cccccc} \text{C\textsubscript{1}} & \text{C\textsubscript{2}} & \text{C\textsubscript{3}} & \text{C\textsubscript{4}} & \text{C\textsubscript{5}} & \text{C\textsubscript{6}}  \\ \text{AVBL} & \text{AIML} & \text{AIBL} & \text{AIAL} & \text{ADAL} & \text{AVAL}\\ \text{AVBR} & \text{AIMR} & \text{AIBR} & \text{AIAR} & \text{ADAR} & \text{AVAR}\\ \text{AVG} & \text{AVFL} & \text{AINL} &   & \text{AVEL} & \text{AVDL}\\ \text{AVJL} & \text{AVFR} & \text{AINR} &   & \text{AVER} & \text{AVDR}\\ \text{AVJR} & \text{AVHL} & \text{AIYL} &   & \text{AVKL} & \text{LUAL}\\ \text{BDUL} & \text{AVHR} & \text{AIYR} &   & \text{AVKR} & \text{LUAR}\\ \text{BDUR} & \text{PVQL} & \text{AIZL} &   & \text{DVA} & \text{PVCL}\\ \text{PVPR} & \text{PVQR} & \text{AIZR} &   & \text{DVC} & \text{PVCR}\\   & \text{RIFL} & \text{AUAL} &   & \text{PVPL} & \text{PVR}\\   & \text{RIFR} & \text{AUAR} &   & \text{PVT} & \text{PVWL}\\   &   & \text{RIAL} &   & \text{RICL} & \text{PVWR}\\   &   & \text{RIAR} &   & \text{RICR} &  \\   &   & \text{RIBL} &   & \text{RIGL} &  \\   &   & \text{RIBR} &   & \text{RIGR} &  \\   &   & \text{RIH} &   & \text{RIPL} &  \\   &   & \text{RIR} &   & \text{RIPR} &  \\   &   &   &   & \text{RIS} &  \\   &   &   &   & \text{RMGL} &  \\   &   &   &   & \text{RMGR} &  \\   &   &   &   & \text{URBL} &  \\   &   &   &   & \text{URBR} &   \end{tabular}}
\end{table}

\begin{table}[ht]
	\caption{\label{tab:motor}Motoneurons of the \textit{C. elegans}. Columns correspond to clusters derived by optimized k-means.}
	\scriptsize
	\centerline{\begin{tabular}{ccccccc} \text{C\textsubscript{1}} & \text{C\textsubscript{2}} & \text{C\textsubscript{3}} & \text{C\textsubscript{4}} & \text{C\textsubscript{5}} & \text{C\textsubscript{6}} & \text{C\textsubscript{7}} \\ \text{AS10} & \text{DD04} & \text{DD05} & \text{DA04} & \text{AVL} & \text{AS11} & \text{AS01}\\ \text{AS06} & \text{VA07} & \text{VA08} & \text{DB03} & \text{DA08} & \text{DA09} & \text{AS02}\\ \text{AS07} & \text{VB06} & \text{VA09} & \text{DB04} & \text{DB07} & \text{DD06} & \text{AS03}\\ \text{AS08} & \text{VB07} & \text{VB08} & \text{DD02} & \text{PVNR} & \text{DVB} & \text{AS04}\\ \text{AS09} & \text{VD07} & \text{VB09} & \text{DD03} & \text{RID} & \text{PDA} & \text{AS05}\\ \text{DA07} & \text{VD08} & \text{VD10} & \text{VA06} & \text{RIML} & \text{PDB} & \text{DA01}\\ \text{DB05} &   & \text{VD09} & \text{VB02} & \text{RIMR} & \text{VA11} & \text{DA02}\\ \text{DB06} &   &   & \text{VB03} & \text{RIVL} & \text{VA12} & \text{DA03}\\ \text{HSNL} &   &   & \text{VB04} & \text{RIVR} & \text{VB10} & \text{DA05}\\ \text{PVNL} &   &   & \text{VB05} & \text{RMDDL} & \text{VB11} & \text{DA06}\\ \text{RMHL} &   &   & \text{VC01} & \text{RMDDR} & \text{VD12} & \text{DB01}\\ \text{RMHR} &   &   & \text{VC02} & \text{RMDL} & \text{VD13} & \text{DB02}\\ \text{SABD} &   &   & \text{VC03} & \text{RMDR} &   & \text{DD01}\\ \text{SABVL} &   &   & \text{VD02} & \text{RMDVL} &   & \text{HSNR}\\ \text{SABVR} &   &   & \text{VD03} & \text{RMDVR} &   & \text{VA01}\\ \text{SIADL} &   &   & \text{VD04} & \text{RMED} &   & \text{VA02}\\ \text{SIADR} &   &   & \text{VD05} & \text{RMEL} &   & \text{VA03}\\ \text{SIAVL} &   &   & \text{VD06} & \text{RMER} &   & \text{VA04}\\ \text{SIAVR} &   &   &   & \text{RMEV} &   & \text{VA05}\\ \text{SIBDL} &   &   &   & \text{RMFL} &   & \text{VB01}\\ \text{SIBDR} &   &   &   & \text{RMFR} &   & \text{VC04}\\ \text{SIBVL} &   &   &   & \text{SAADL} &   & \text{VD01}\\ \text{SIBVR} &   &   &   & \text{SAADR} &   &  \\ \text{SMBDR} &   &   &   & \text{SAAVL} &   &  \\ \text{URADL} &   &   &   & \text{SAAVR} &   &  \\ \text{VA10} &   &   &   & \text{SMBDL} &   &  \\ \text{VC05} &   &   &   & \text{SMBVL} &   &  \\   &   &   &   & \text{SMBVR} &   &  \\   &   &   &   & \text{SMDDL} &   &  \\   &   &   &   & \text{SMDDR} &   &  \\   &   &   &   & \text{SMDVL} &   &  \\   &   &   &   & \text{SMDVR} &   &  \\   &   &   &   & \text{URADR} &   &  \\   &   &   &   & \text{URAVL} &   &  \\   &   &   &   & \text{URAVR} &   &  \\   &   &   &   & \text{VD11} &   &   \end{tabular}}
\end{table}

\subsection{Evaluation of The Clustering}

In order to evaluate the inspected clusters of sensory neurons (Fig. \ref{fig:sensory}B), interneurons (Fig. \ref{fig:inter}B) and motoneurons (Fig. \ref{fig:motor}B), we used statistical testing of communities (clusters). In all three cases, the nodes with importance $m_i=0$ are considered as one additional cluster. We use the Newman-Girvan modularity as statistic~\citep{newman2006}. A vector of nodal assignments to clusters expresses its goodness of fit to the underlying adjacency matrix through the value of modularity $Q$. It is calculated as:

\begin{equation}
Q=\frac{1}{2w}\sum_{i,j}^N([\mathbf{A}_{bin}]_{i,j}-\frac{d_id_j}{2w})\delta_{C_i, C_j},
\label{eq:modularity}
\end{equation}

\noindent where $N$ is the number of nodes, $w$ is the total strength of edges in the graph, $\ma A_{bin}$ is the graph binary adjacency matrix, $d_i$ denotes the degree of the $i$\textsuperscript{th} node, $\delta$ is the Kronecker delta function, and $C_i$ denotes the cluster to which the $i$\textsuperscript{th} node belongs.


In Supplementary Fig. \ref{fig:modularity}, we present the results of the statistical approach for the case of sensory (red plots), inter- (grey plots) and motoneurons (green plots). The modularity values for the assignments to clusters as found by k-means clustering (see Section~\ref{subsec::clustering}) are marked with the dashed lines and labeled with $Q\textsubscript{sensory}$, $Q\textsubscript{inter-}$, and $Q\textsubscript{moto-}$ (Supplementary Fig.~\ref{fig:modularity}B). For the number of clusters estimated by the Silhouette method, we generated 999 random assignment vectors and calculated $Q$ each time, in order to build a null distribution (Supplementary Fig.~\ref{fig:modularity}A).

As $Q\textsubscript{sensory}$, $Q\textsubscript{inter-}$, and $Q\textsubscript{moto-}$ are above the corresponding distributions of modularity for random assignments, we conclude that the found clustering is significant. Since these modularity values are strictly greater than all other $Q$ values for random assignments, and, consequently, from any chosen percentile of the calculated distributions, the test rejects the null hypothesis that the chosen clustering is random at even very small significance levels. Finally, we note that the distribution of $Q$ in the case of interneurons is slightly closer to  the corresponding value of $Q\textsubscript{inter-}$ than in the case of sensory or motoneurons. This can be expected, since interneurons are more strongly connected to other cell types, and thus, do not impose as strong communities as for the clusters formed from sensory or motoneurons.

\subsection{Supplementary Figures}

\renewcommand{\thefigure}{S\arabic{figure}}
\setcounter{figure}{0}

\begin{figure}
\centering
\includegraphics[height=0.65\textheight]{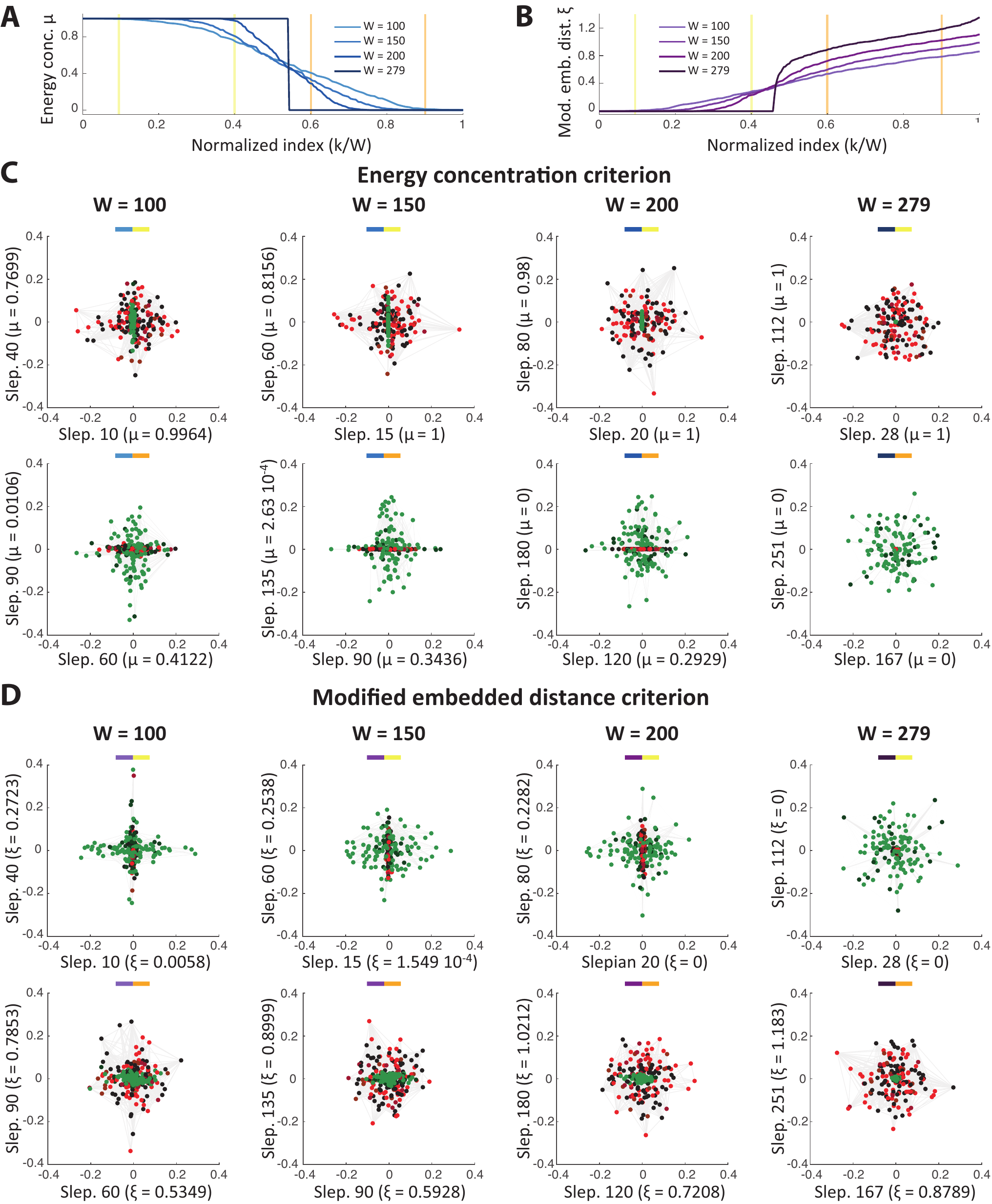}
\caption{ For energy concentration (\textbf{A}) and modified embedded distance (\textbf{B}) criteria, eigenspectra at bandwidth $W=100,150,200,279$, as depicted by increasingly darker blue or purple shades, respectively. Yellow and orange vertical bars map locations of the eigenspectra at which Slepian vectors are shown (\textbf{C} and \textbf{D}). They are displayed for increasing bandwidth going from left ($W=100$) to right ($W=279$, full bandwidth). For energy concentration (\textbf{C}), the first row illustrates two Slepian vectors mapping the start of the spectrum (normalized indices of $0.1$ --- strongly concentrated in $\mathcal{S}$ --- and $0.4$ --- still concentrated, but less for lower bandwidth). The second row denotes two Slepian vectors from the second half of the spectrum (normalized indices of $0.6$ --- mildly concentrated in $\mathcal{S}$ using a smaller bandwidth --- and $0.9$ --- not concentrated at all). Visualizations are similar for modified embedded distance (\textbf{D}), but in this case, low eigenvalues imply either non-concentrated (\textit{e.g.}, X axis, first row of plots) or mildly concentrated but low localized spatial frequency Slepian vectors (for instance, Y axis, first row of plots, $W=200$), while high eigenvalues relate to high localized spatial frequency Slepian vectors (see Y axis, second row of plots). See \cite{VDV2017_Guiding} for another preliminary analysis of the dataset from the modified embedded distance viewpoint. $\mu$ and $\xi$ values of the shown Slepian vectors are provided in parentheses on each axis.}
\label{fig:sf1}
\end{figure}

\begin{figure}
\centering
\includegraphics[width=0.85\textwidth]{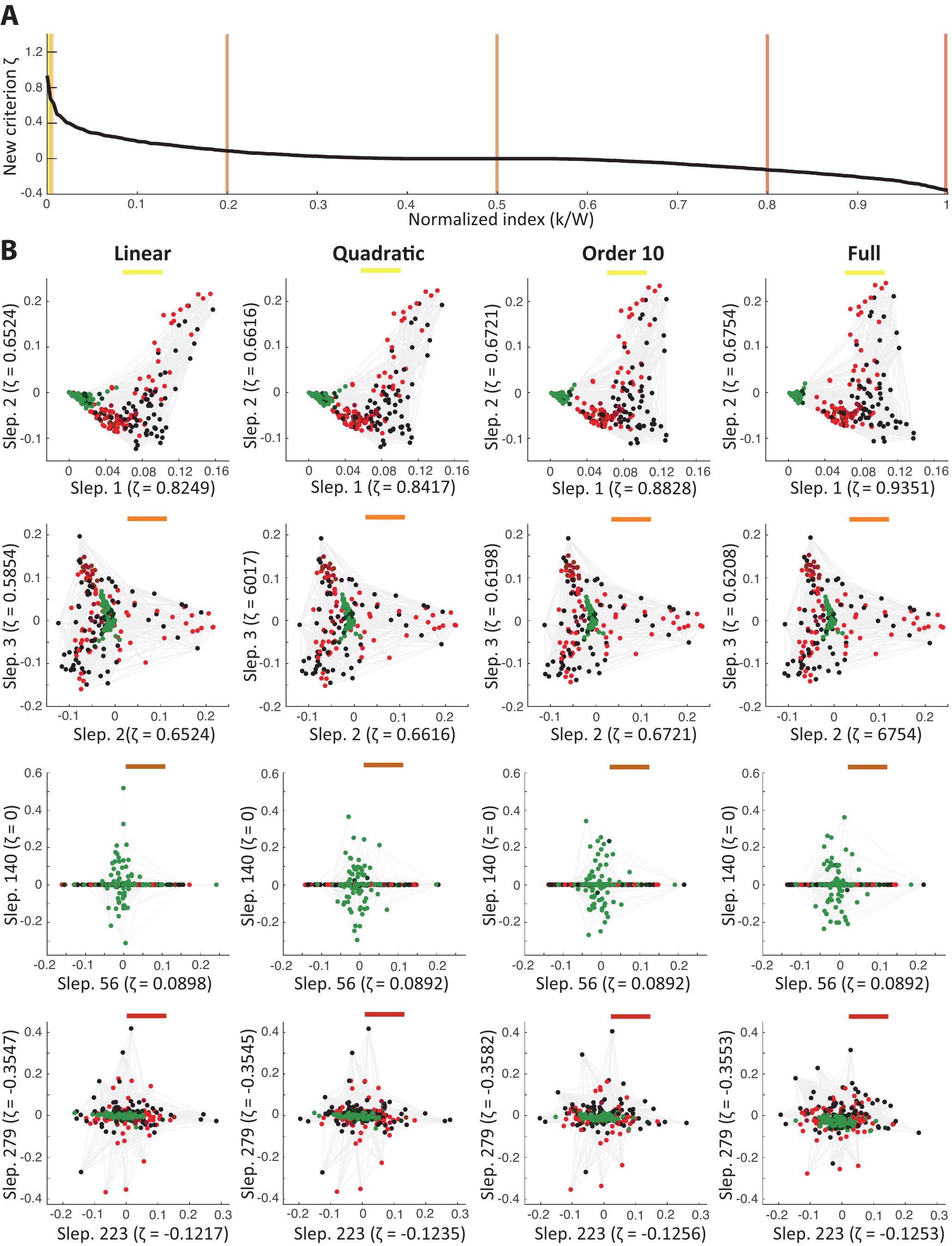}
\caption{\label{fig:sf2} Eigenspectrum of the newly developped $\zeta$ criterion (\textbf{A}), with vertical bars highlighting the locations of the spectrum at which four pairs of Slepian vectors were sampled for display (\textbf{B}, from first to fourth row as respectively depicted by yellow, orange, brown and red color codes). Results obtained with linear, quadratic and order 10 approximations, as well as from a full computation of $\mathbf{M}-\mathbf{L}^{1/2}\mathbf{M}\mathbf{L}^{1/2}$, are respectively shown from left to right. $\zeta$ values of the shown Slepian vectors are provided in parentheses on each axis.}
\end{figure}

\begin{figure}
\centering
\includegraphics[width=0.9\textwidth]{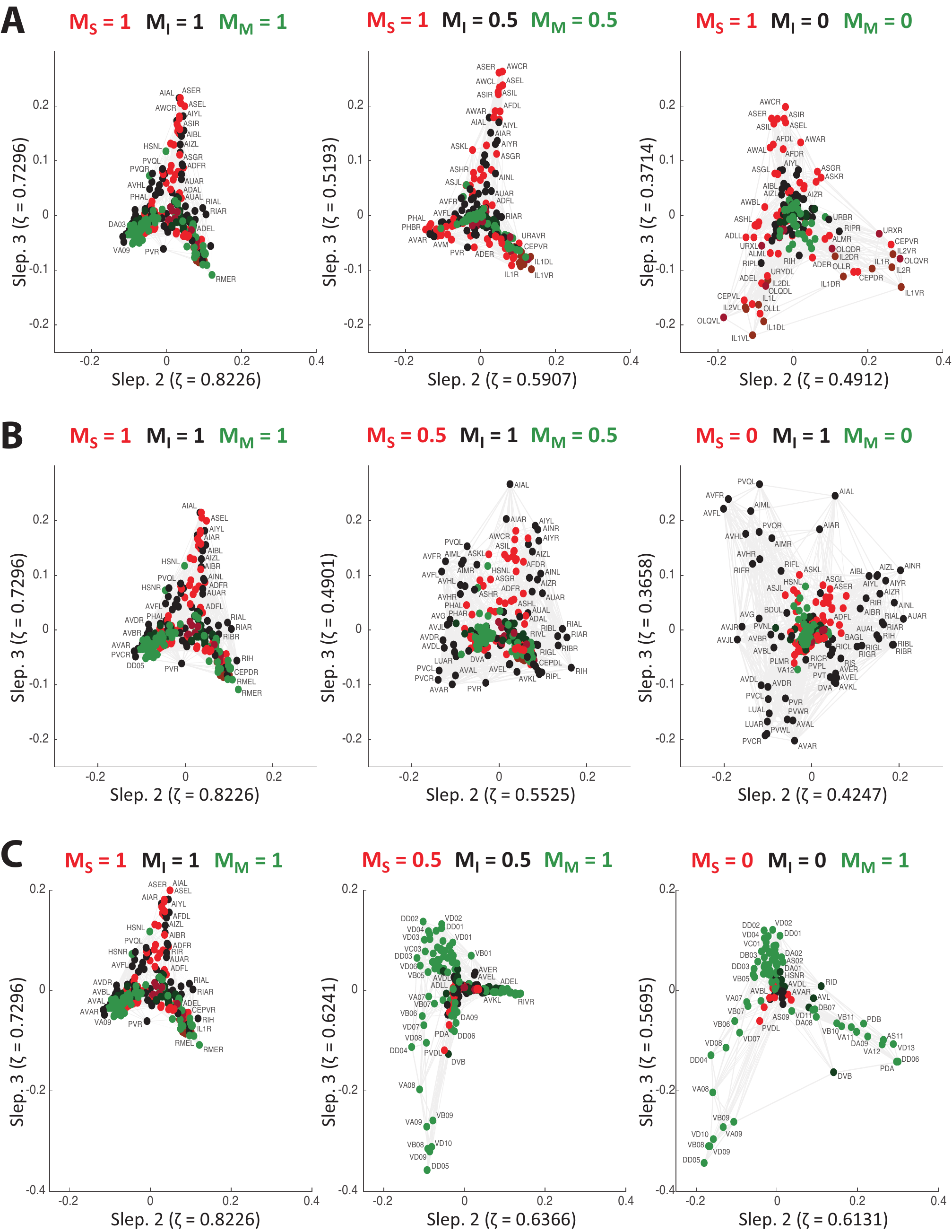}
\caption{\label{fig:sf3} Start (left column), intermediate (middle column) and end (right column) representations of sensory neuron (\textbf{A}), interneuron (\textbf{B}) or motoneuron (\textbf{C}) trajectories, respectively, setting cooperation weights for other neuron types to $1$, $0.5$ or $0$. Cells are labeled according to \cite{Varshney2011}. The start representation is the same across cases, since then $\mathbf{M} = \mathbf{I}$ and the problem boils down to the eigendecomposition of the adjacency matrix $\mathbf{A}$, or equivalently of the Laplacian $\mathbf{L}=\mathbf{I}-\mathbf{A}$ highlighted in Fig.~\ref{fig:principle}.}
\end{figure}

\begin{figure}
	\centering
	\includegraphics[width=0.95\textwidth]{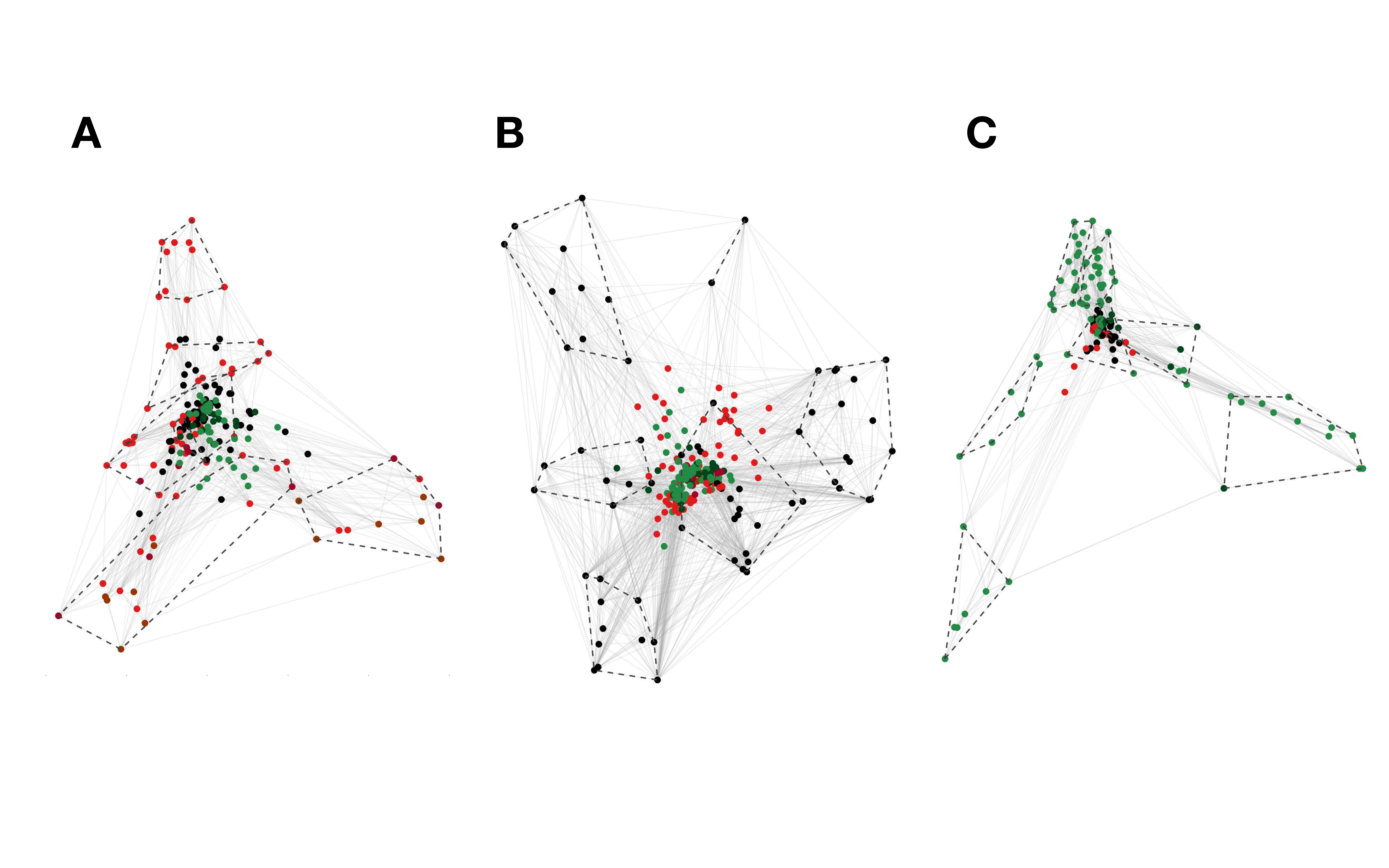}
	\caption{\label{fig:clusters} Clusters derived by repeated k-means clustering of the focused nodes in the case of sensory neurons (\textbf{A}), interneurons (\textbf{B}) or motoneurons (\textbf{C}). The optimal number of clusters was estimated with the Silhouette approach. Nodes constituting the border of each cluster's convex hull are connected by a dashed black line to visualize the clusters.}
\end{figure}

\begin{figure}
\centering
\includegraphics[width=0.85\textwidth]{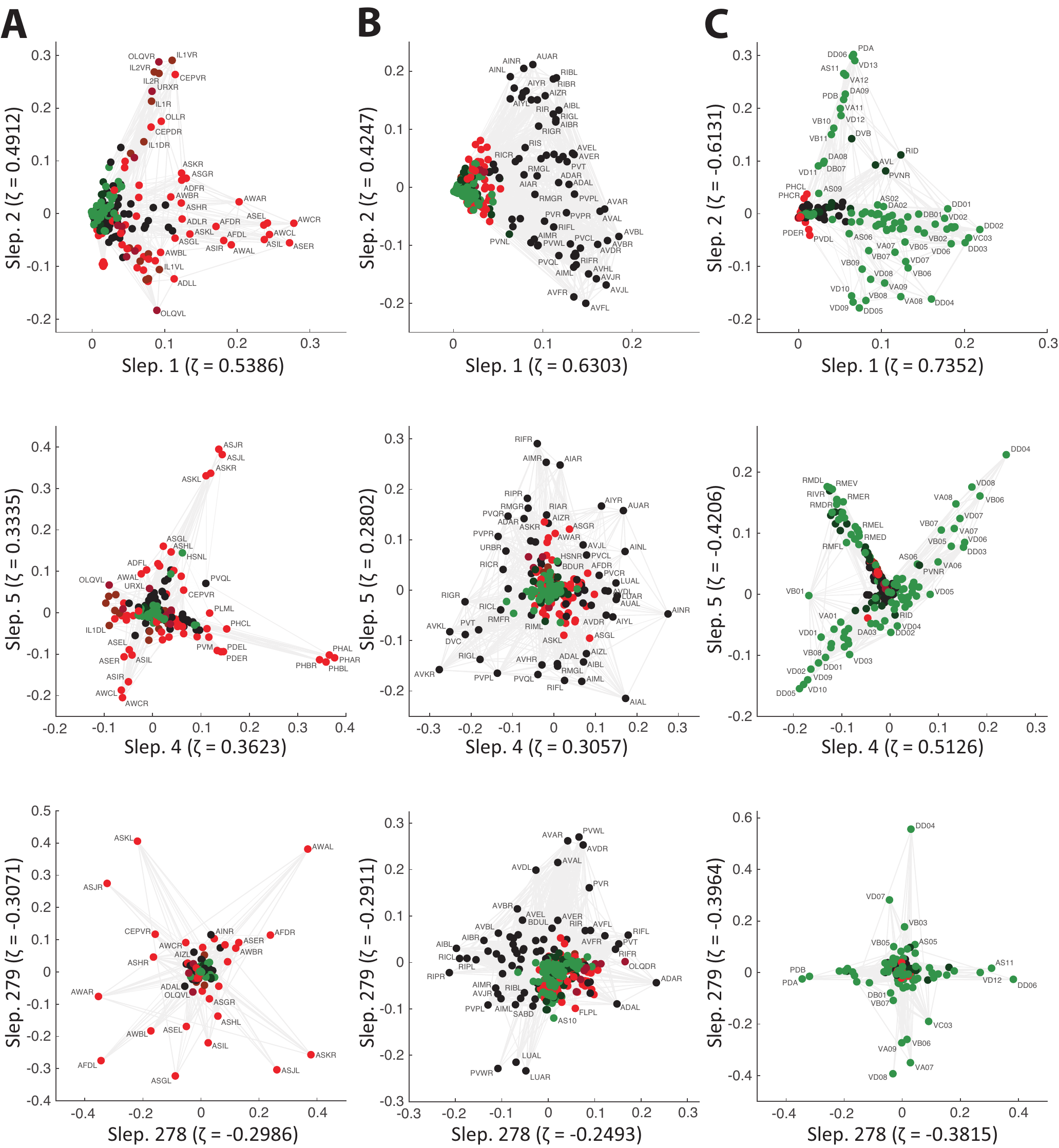}
\caption{\label{fig:sf4} Focussing on (\textbf{A}) sensory neurons (red), (\textbf{B}) interneurons (black) or (\textbf{C}) motoneurons (green), two-dimensional visualization using alternative sets of Slepian vectors: first and second (first row), fourth and fifth (second row), or last two (third row). Cells are labeled according to \cite{Varshney2011}. $\zeta$ values of the shown Slepian vectors are provided in parentheses on each axis.}
\end{figure}

\begin{figure}
\centering
\includegraphics[width=0.80\textwidth]{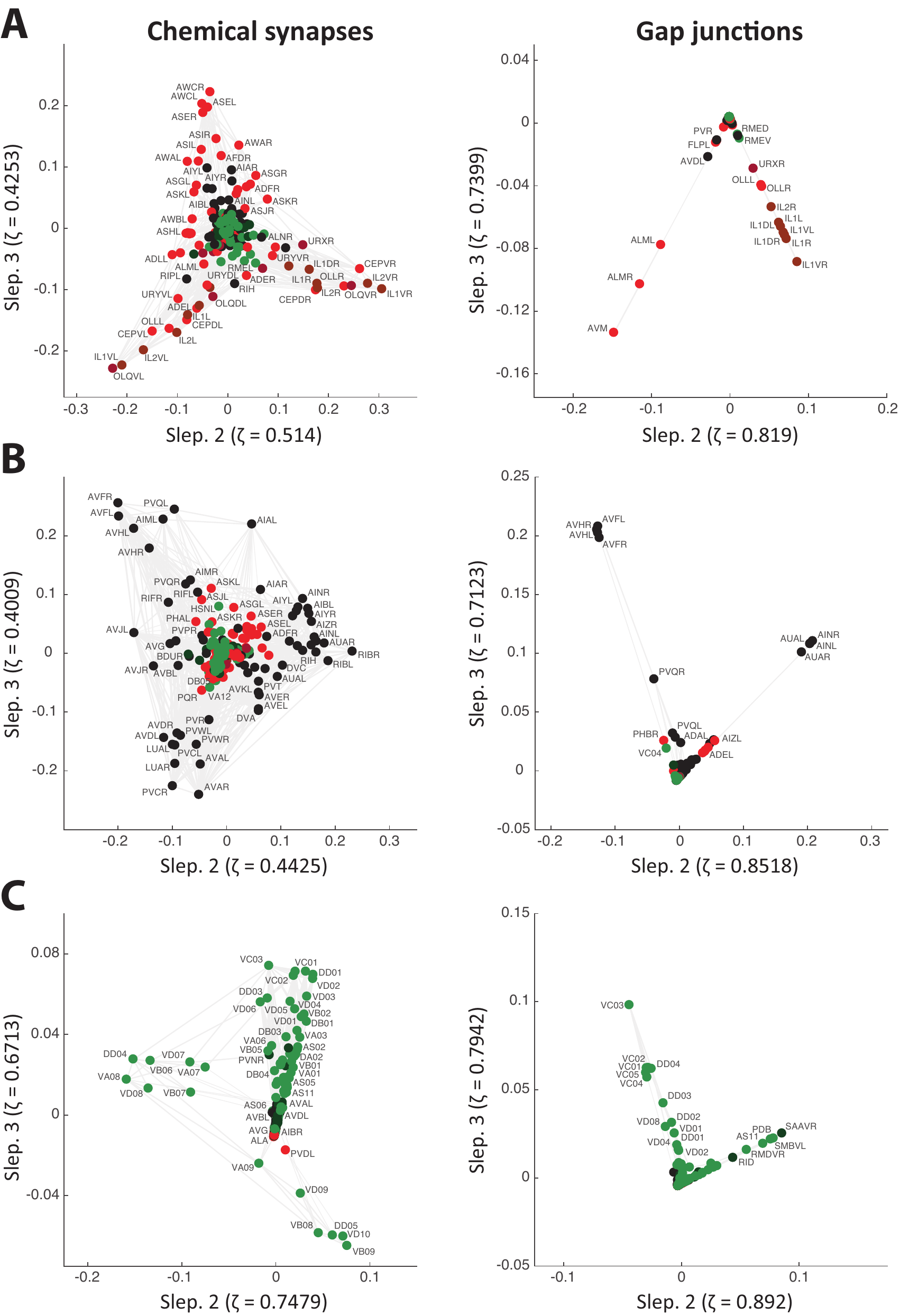}
\caption{\label{fig:sf5} Separate two-dimensional visualizations when only considering chemical synapses (left column) or gap junctions (right column) for sensory neurons (\textbf{A}), interneurons (\textbf{B}) or motoneurons (\textbf{C}). Cells are labeled according to \cite{Varshney2011}. $\zeta$ values of the shown Slepian vectors are provided in parentheses on each axis.}
\end{figure}

\begin{figure}
\centering
\includegraphics[width=0.95\textwidth]{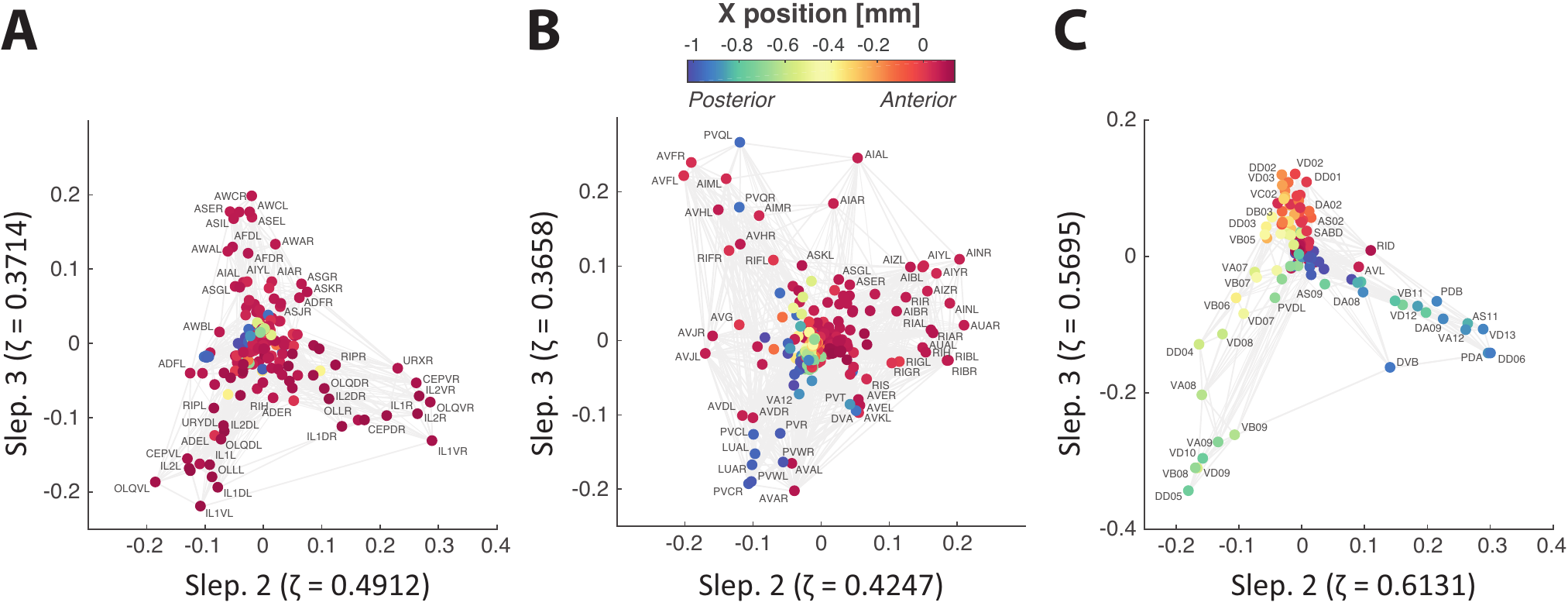}
\caption{\label{fig:sf6} Two-dimensional visualizations for sensory neurons (\textbf{A}), interneurons (\textbf{B}) or motoneurons (\textbf{C}) when representing each neuron as a function of its position along the X direction. A value of 0 indicates the location of the nerve ring, and positional data was retrieved from \cite{Varier2011}. $\zeta$ values of the shown Slepian vectors are provided in parentheses on each axis.}
\end{figure}


\begin{figure}
	\centering
	\includegraphics[width=0.85\textwidth]{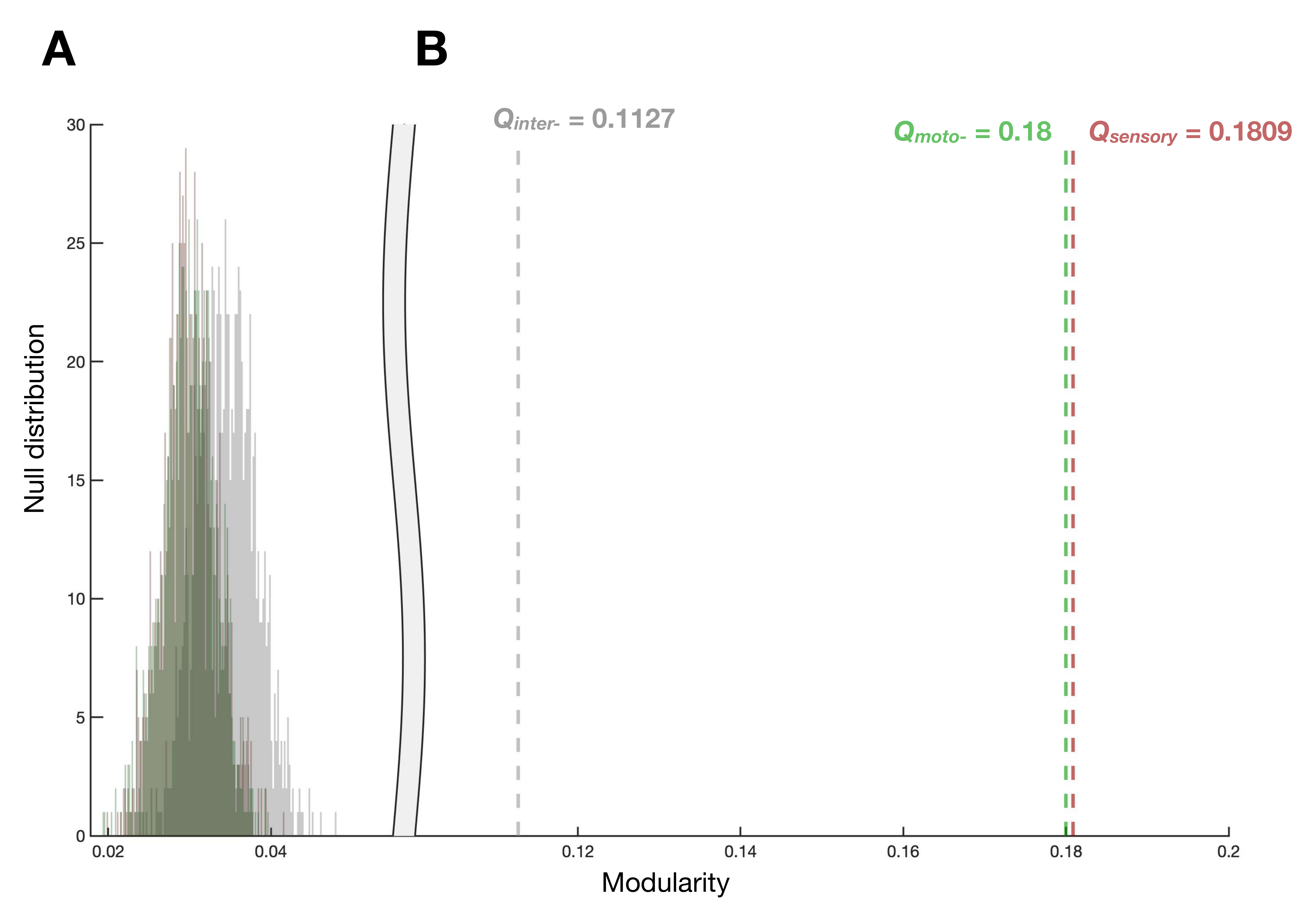}
	\caption{\label{fig:modularity} Testing of clustering assignments when the focus is on sensory neurons (red), interneurons (grey) or motoneurons (green). For each case, all nodes of other types are considered as one additional cluster. The null distributions of the modularity of random assignments to clusters are given on the left side of the plot (\textbf{A}).The dashed straight lines on the right represent values of modularity $Q$ for the clusters in Fig. \ref{fig:clusters} derived by the k-means approach (\textbf{B}). The x-axis is broken at $0.06$ for better visualization.}
\end{figure}

\newpage

\clearpage
\bibliography{NETN}

\end{document}